\documentclass[journal]{IEEEtran}

\usepackage{cite}
\usepackage{bm}

\usepackage{subfigure}
\usepackage{booktabs} 
\usepackage{multirow} 
\usepackage{times}
\usepackage{graphicx} %
\usepackage{color}

\ifCLASSINFOpdf
\else

\fi
\usepackage{amsmath}
\usepackage{algorithmic}
\usepackage{algorithm}
\usepackage{amsfonts}
\usepackage{amsthm}
\usepackage{mathtools}

\newcommand{\SUB}[1]{\ENSURE \hspace{-0.15in} \textbf{#1}}
\newcommand{\nc}{K}

\begin{document}

\title{Towards Tailored Models on Private AIoT Devices: Federated Direct Neural Architecture Search}

\author{Chunhui~Zhang, Xiaoming Yuan, Qianyun Zhang, Guangxu Zhu, Lei Cheng, and Ning Zhang
\thanks{The work was developed during C. Zhang’s visit to L. Cheng's research group at Zhejiang University.}
\thanks{The work was  supported in part by the National Natural Science Foundation of China under Grant 62001309,  62001310, 61901020, and 61901099; in part by the National Key Research and Development Program of China under Grant 2018YFB1800800; and in part by  the Natural Science Foundation of Hebei Province under Grant F2020501037. {\it (Corresponding author:  Lei Cheng.)}
}
\thanks{C. Zhang is with the Michtom School of Computer Science, Brandeis University, Waltham, MA, USA (email: chunhuizhang@brandeis.edu).}
\thanks{X. Yuan is with the Qinhuangdao Branch Campus, Northeastern University, Qinhuangdao, 066004, China (Email:yuanxiaoming@neuq.edu.cn).}
\thanks{Q. Zhang  is with the School of Cyber Science and Technology, Beihang University, Beijing, China (email: zhangqianyun@buaa.edu.cn).}
\thanks{G. Zhu is with the Shenzhen Research Institute of Big Data, Shenzhen, Guangdong, P. R. China (e-mail: gxzhu@sribd.cn).}
\thanks{L. Cheng is with the College of Information Science and Electronic Engineering, Zhejiang University, and also with 
 Zhejiang Provincial Key Laboratory of Info. Proc., Commun. \& Netw. (IPCAN), Hangzhou, 310027, China (email: lei\_cheng@zju.edu.cn).}
\thanks{N. Zhang is with the Department of Electrical and Computing Engineering, University of Windsor, Windsor, ON N9B 3P4, Canada (e-mail:
ning.zhang@uwindsor.ca).}
}

\markboth{}%
{Shell \MakeLowercase{\textit{et al.}}: Bare Demo of IEEEtran.cls for IEEE Journals}

\maketitle

\begin{abstract}
Neural networks often encounter various stringent resource constraints while deploying on edge devices. To tackle these problems with less human efforts, automated machine learning becomes popular in finding various neural architectures that fit diverse Artificial Intelligence of Things (AIoT) scenarios.
Recently, to prevent the leakage of private information while enable automated machine intelligence, there is an emerging trend to integrate federated learning and neural architecture search (NAS). Although promising as it may seem, the coupling of difficulties from both tenets makes the algorithm development quite challenging. In particular, how to efficiently search the optimal neural architecture directly from massive non-independent and identically distributed (non-IID) data among AIoT devices in a federated manner is a hard nut to crack. In this paper, to tackle this challenge, by leveraging the advances in ProxylessNAS, we propose a Federated Direct Neural Architecture Search (FDNAS) framework that allows for hardware-friendly NAS from non-IID data across devices. To further adapt to both various data distributions and different type of devices with heterogeneous embedded hardware platforms, inspired by meta-learning, a Cluster Federated Direct Neural Architecture Search (CFDNAS) framework is proposed to achieve device-aware NAS, in the sense that each device can learn a tailored deep learning model for its particular data distribution and hardware constraint. Extensive experiments on non-IID datasets have shown the state-of-the-art accuracy-efficiency trade-offs achieved by the proposed solution in the presence of both data and device heterogeneity. 
\end{abstract}

\IEEEpeerreviewmaketitle
\section{Introduction}

\IEEEPARstart{F}{rom} AlexNet \cite{krizhevsky2012imagenet}, VGG \cite{simonyan2014very}, ResNet \cite{he2016deep}, to SENet \cite{hu2018squeeze}, tremendous research efforts have been put in efficient deep learning model designs, leading to state-of-the-art performances for various machine learning tasks in the context of Artificial Intelligence of Things (AIoT), including image classification in mobile phones \cite{howard2019searching}, object detection in self-driving automobiles \cite{redmon2018yolov3}, and medical image segmentation in computer aided diagnosis \cite{ren2015faster, long2015fully}. 

In AIoT, intelligent edge devices (also called AIoT devices) are usually memory-constrained. For instance, the inference memory of the popular single-board computer \textit{Raspberry Pi 1 Model A} is only 256 MB \cite{ji2020mcunet}, which is too small to allow the deployment of large deep learning models (e.g., VGG-16 consumes about 14 GB memory). In addition, in many scenarios, AIoT devices are required to perform real-time inference~\cite{8693826}. For example, both face recognition on mobile phones and real-time objection detection on unmanned aerial vehicles (UAV) demand a low latency of fewer than 500 milliseconds \cite{wang2020scaled}. Therefore, the light-weight deep learning models are highly desired for \textit{Artificial Intelligence of Things (AIoT)} devices.

To achieve remarkable inference performance on light-weight deep learning models deployed on AIoT devices, adequate training using a vast amount of data is imperative, which however faces the practical challenge of big data collection at the server, due to the data privacy concern that has received increasing attentions recently~\cite{pmlr-v54-mcmahan17a}. 

To attain a fully trained model while protecting AIoT devices' data privacy, federated learning has been proposed recently as a promising solution~\cite{pmlr-v54-mcmahan17a}. Instead of training models in a centralized manner with direct access to the raw data from devices, federated learning requires no data access at the server. In particular, each device trains the machine learning model on its local machine and then uploads the model parameters to the central server. Since only learning parameters are exchanged in the air, the risk of data leakage is largely decreased~\cite{truex2019hybrid}. Although straightforward as it may seem, integrating training methods developed from classic centralized machine learning into the framework of federated learning with indistinguishable performance loss is not a simple matter. It has triggered increasing research efforts in both machine learning and optimization societies~\cite{li2020on}. 

For various AIoT devices, to learn their own light-weight deep neural architectures in a federated manner, it is crucial to take the non-independent and identically distributed (non-IID) nature of the data across devices into account. 
Thanks to gradient-based neural architecture search (NAS) framework \cite{liu2018darts, cai2018proxylessnas}, the model search can be performed by updating the associated architecture parameters. 
 {\color{black}Some NAS methods search for backbone cells (i.e., the main structure of a deep learning model) on proxy data (i.e., a separate small dataset that is similar to the target data) to trim down computational cost, such as DARTS \cite{liu2018darts}, PC-DARTS \cite{ xu2020pcdarts}, and MiLeNAS \cite{he2020milenas}. However, these proxy-based strategies do not guarantee that the searched backbone cells can maintain the similar excellent performance on the target data\cite{yang2019evaluation}. Also, these methods require practitioners to decide the appropriate depth of the cell block stacking for different scenarios. This issue will be amplified in the framework of federated learning since the discrepancy between proxy data and target data will be enlarged due to the data heterogeneity across devices~\cite{8889996}. }
Therefore, under the framework of federated learning, there is a need to seek a ProxylessNAS method that {\it directly learns light-weight neural networks from the target data among devices}. We term this scheme as Federated Direct Neural Architecture Search (\textbf{FDNAS}). 

On the other hand, according to their hardware platforms and data distribution, AIoT devices can be divided into different groups, each containing those devices with similar {\it data distributions} and {\it hardware platforms}.
{\color{black} Therefore, instead of looking for a common neural architecture for all the devices, we expect that FDNAS should exploit the group similarities among devices while allow each device to have its personalized neural architecture tailored to its own task and hardware platform. }
To achieve this goal, we assume that the models of different devices can form a large ensemble, which contains multiple \textit{highly diverse} and \textit{device-specific} models as its subsets. More concretely, this ensemble can be viewed as a SuperNet, and the deep learning model of each device is regarded as a SubNet. All the weights of SubNets are coupled together. 
Then, a \textbf{primary question} is: \emph{ 
how can different devices collaboratively search for multiple tailored models that are near-optimal on their own specific data or hardware, while their weights are entangled in one unified ensemble? 
}
To find the diverse tailored models in such a complex ensemble, we resort to the fundamental idea of {\it meta-learning}. 

Notably, in meta-learning, weights of a deep learning model that have been already meta-trained can have a high transferability when they are adapted to different tasks via meta-test \cite{chen2019closerfewshot}.
Therefore, SuperNet is proposed to be used in a meta-test-like manner to obtain all device-specific neural architectures in federated learning. In the order of ``meta-train" to ``meta-test", the SuperNet trained on all devices in the first phase are considered as the meta-training model for the next ``meta-test" device-specific adaptation. Following this idea, we propose the Cluster Federated Direct Neural Architecture Search (\textbf{CFDNAS}) that divides all the devices into groups by their data similarities and hardware platforms, and each group is trained using the SuperNet from the previous phase. In turn, each group utilizes the previous SuperNet and can adjust the architecture to fit their own device data after only a few rounds of updates (similar to a meta-test). Consequently, CFDNAS can quickly generate a personalized neural architecture for each device in parallel, under the framework of federated learning.

The contributions of this paper are summarized as follows: 
\begin{itemize}
\item[1.] 
We integrated federated learning with gradient-based ProxylessNAS. This strategy allows federated learning not only to train model weights but also to search model architectures directly from the data and hardware latency tables in AIoT devices. {\color{black} Different from previous work \cite{he2020fednas} that employed MiLeNAS \cite{he2020milenas}, which demands the proxy data and performs the cell-level NAS, this paper exemplifies the development of proxyless and layer-level NAS in the context of federated learning, which has the potential to give the searched neural network with enhanced classification performance.}

\item[2.] 
We extended FDNAS to CFDNAS for better exploiting the model diversity. Instead of training a common deep learning model for diverse AIoT models, the exploitation of model diversity enables each device to learn its tailored deep learning model. 

\item[3.]
Inspired by meta-learning strategy, CFDNAS learns multiple device-specific models in parallel, which significantly accelerates the training process.

\item[4.]
{\color{black} Extensive experiment results are reported using various datasets, including CIFAR-10 \cite{cifar10},  ImageNet \cite{5206848}, LEAF \cite{caldas2018leaf}, and FedML \cite{he2020fedml},  to showcase the excellent performance of the proposed scheme. To measure  the performance of the  proposed scheme in real-world AIoT devices, we include the experiment results on the ultrabook-level laptop and mobile phone (SAMSUNG GALAXY S20). }

\end{itemize}




The rest of this paper is organized as follows. In Section \ref{sec:related}, we review the related works of efficient neural  architecture design and federated learning, in which the unique features of the proposed FDNAS and CFDNAS are pinpointed by comparing with the prior works. In Section \ref{sec:preliminary}, we introduce the preliminaries of federated learning and efficient neural architecture search: Federated Averaging and ProxylessNAS, based on which the FDNAS and CFDNAS are {\color{black} developed}  in Section \ref{sec:fdnas}. 
Section \ref{sec:exp} presents experimental results using real-world dataset to show the excellent performance of the proposed algorithms. Finally, concluding remarks and future research are discussed in Section \ref{sec:con}.

\section{Related Works and Background}
\label{sec:related}
\textbf{Efficient neural architecture design} is an active research area due to its importance in supporting the vision of AIoT.
MobileNetV2 \cite{sandler2018mobilenetv2} introduced MBconv blocks, which largely reduce the model's floating-point operations (FLOPs) to save the computations of mobile phones. 
To speed up the NAS, the one-shot NAS approach~\cite{brock2018smash} treats all normal nets as different SubNets of the SuperNet and shares weights among the operation candidates.
ENAS \cite{pham2018efficient} uses the RNN controller to sample SubNets in SuperNet and uses reinforce method to obtain approximate gradients of architecture.
DARTS proposed in~\cite{liu2018darts} improves the search efficiency by representing each edge as a mixture of candidate operations and optimizing the weights of operation candidates via continuous relaxations.
More recently, hardware-friendly NAS methods like ProxylessNAS incorporate latency feedback into the joint optimization task \cite{wu2019fbnet, wan2020fbnetv2, cai2018proxylessnas}.
In parallel, some NAS methods search for the best backbone cells and transfer them to other target tasks by stacking them together \cite{liu2018darts, xu2020pcdarts}. {\color{black} Recently,  MiLeNAS uses gumble-softmax for one-round searching to enhance the speed of SuperNet optimization\cite{he2020milenas}.}
However, due to the difference between the proxy data and the target data, the best block searched by the proxy method differs from the optimal normal net~\cite{yang2019evaluation}.
Prior works \cite{redmon2018yolov3, cai2018proxylessnas} did not take the privacy issue of devices into account and thus has not connected to the framework of federated learning. In contrast, this paper aims at designing a privacy-preserving learning scheme for AIoT devices. Furthermore, the proposed scheme is devised to learn device-specific models tailored to their specific hardware platforms and data distribution, without any proxy-based pre-training step. To the best knowledge of authors, there is no such design that enjoys the above merits simultaneously, motivating the current works.

\textbf{Federated learning} is a recently proposed privacy-preserving distributed learning framework, where the neural network architecture is pre-determined, and the learning process is only to optimize the model weights. FedAvg is one of the most widely adopted training algorithms~\cite{pmlr-v54-mcmahan17a}, where local model instead of raw data is exchanged between the server and devices to avoid private data leakage.
There are some fundamental challenges in federated learning that attract increasing research interests~(see~\cite{li2019federated} and references therein), including communication overhead, statistical heterogeneity of data (i.e., non-IID nature), and data privacy.
Particularly, the communication between the central server and the devices becomes the bottleneck due to the frequent exchange of high-dimensional weights. This motivates recent studies that aim at designing more communication-efficient strategies \cite{konevcny2016federated, 45672, fedpaq19,9272666}.
In practice, the data is usually non-IID across devices. This motivates the investigation of hyper-parameter settings of FedAvg (e.g., learning rate decay) to demystify their impacts on non-IID data \cite{li2020on}. In addition, a global data-sharing strategy is proposed to improve the accuracy of the algorithm on non-IID data \cite{zhao2018federated}.
On the other hand, privacy issue was also the focus of recent studies \cite{agarwal2018cpsgd}. For example, differential privacy is applied to federated learning for further protection of data privacy.  \cite{article17eth, 9069945}. Most recently, researchers started to look into the possible integration of federated learning and NAS. 
{\color{black}Prior work FedNAS~\cite{he2020fednas}, using MiLeNAS for search efficiency~\cite{he2020milenas}, first searched the cell architecture as the proxy task and then transferred the learnt structure to the target dataset.
Note that all of the above techniques either requires pre-defined network architectures or search for backbone cells using proxy strategies. In contrast, the proposed FDNAS and CFDNAS can not only search the complete neural architecture directly from the data at devices but also allow each device to learn a personalized deep learning model in a federated manner. Notice that for CFDNAS, multiple suitable neural networks are learnt in parallel without expensive training costs.}

\section{Preliminary}
\label{sec:preliminary}
In this section, we briefly review the basics of federated learning and ProxylessNAS. 
\subsection{Federated Averaging}
\label{sec:fedavg}

As an emerging privacy-protection technology, federated learning enables edge devices to collaboratively train a shared global model without uploading their private data to a central server~\cite{pmlr-v54-mcmahan17a}. To this end, at training round $t+1$, each edge device  $k\in S$ downloads the shared machine learning model $\mathbf w_{t}$ (e.g., a CNN model) from the central server and utilizes its local data to update the local model weights. Then, each device sends their locally updated models $\{\mathbf w^{k}_{t} \}_{k \in S}$ to the central server for aggregation. This procedure resembles conventional distributed machine learning on parallel computational architectures (e.g., parameter server). However, there is a significant difference.  Without data collection and reshuffling, training data in each local device is usually non-IID, which would hamper the efficient training of the machine learning model. To allow fully training, frequent model weights exchange between devices and server is needed, which would cause heavy communication overhead.

To tackle the challenges, in federated averaging (FedAvg) algorithm~\cite{pmlr-v54-mcmahan17a}, as summarized in {\bf Algorithm~\ref{alg:fedavg}}, has been proposed recently to dramatically reduce communication rounds by simultaneously increasing local training epochs and decreasing mini-batch sizes in local stochastic gradient descent (SGD) steps. Furthermore, recent theoretical research~\cite{li2020on} reveals that the global convergence of FedAvg algorithm can be linked to the differences of data distributions among edge devices, giving a glimmer of hope for approaching the global optimum with a theoretical guarantee. 

\begin{algorithm}[t]\small
    \caption{Federated Averaging \cite{pmlr-v54-mcmahan17a}: $K$ denotes the number of devices; $T$ is the number of communication rounds; $\eta$ denotes the learning rate; $N_k$ is the size of local training dataset in device $k$; $\mathbf w_0^{g}$ denotes server's global weight parameters; and $N$ is the size of the total training dataset.}
    \label{alg:fedavg}
    \begin{algorithmic} 
    \SUB{Central server:}
  \STATE Initialize $\mathbf w_0^g$
  \FOR{each communication round $t = 1, 2, \dots T$}
     \FOR{each device $k \in S$ \textbf{in parallel}}
      \STATE $\mathbf w_{t+1}^k \leftarrow \text{DeviceUpdate}_k(\mathbf w_t^g)$ 
     \ENDFOR
     \STATE $\mathbf w_{t+1}^g \leftarrow \sum_{k=1}^\nc \frac{N_k}{N} \mathbf w_{t+1}^k$ 
  \ENDFOR
  \STATE

\SUB{DeviceUpdate($\mathbf w$):} // on device platform.
  \FOR{each epoch $\mathbf e$}
    \FOR{ each mini-batch $\mathbf b$ }
      \STATE $\mathbf w \leftarrow \mathbf w - \eta  \nabla \mathcal{L}(\mathbf w, \mathbf b)$
    \ENDFOR
 \ENDFOR
 \STATE return $\mathbf w$ to server
    \end{algorithmic}
\end{algorithm}

\subsection{ProxylessNAS}
\label{sec:plnas}
The surging interest in NAS has been recently prompted by its great success in automating neural network design, which has outperformed many manually designed counterparts in various deep learning tasks. However, most NAS methods need prohibitively intensive computational resources to achieve the best performance, hindering the wide adoption in practice. To overcome this hurdle, the framework of ProxylessNAS emerged recently that can save computational resources significantly while at nearly no cost of performance loss.  As a result,  it opens the door to directly learn the optimal architecture from large-scale datasets without resorting to a proxy-based scheme. Furthermore, it allows the incorporation of inference latency on hardware platforms into the neural network architecture design, thus paving the way to speed up the inference for customized AIoT devices.

\begin{figure}[t]
    \centering
    \includegraphics[width=1\linewidth]{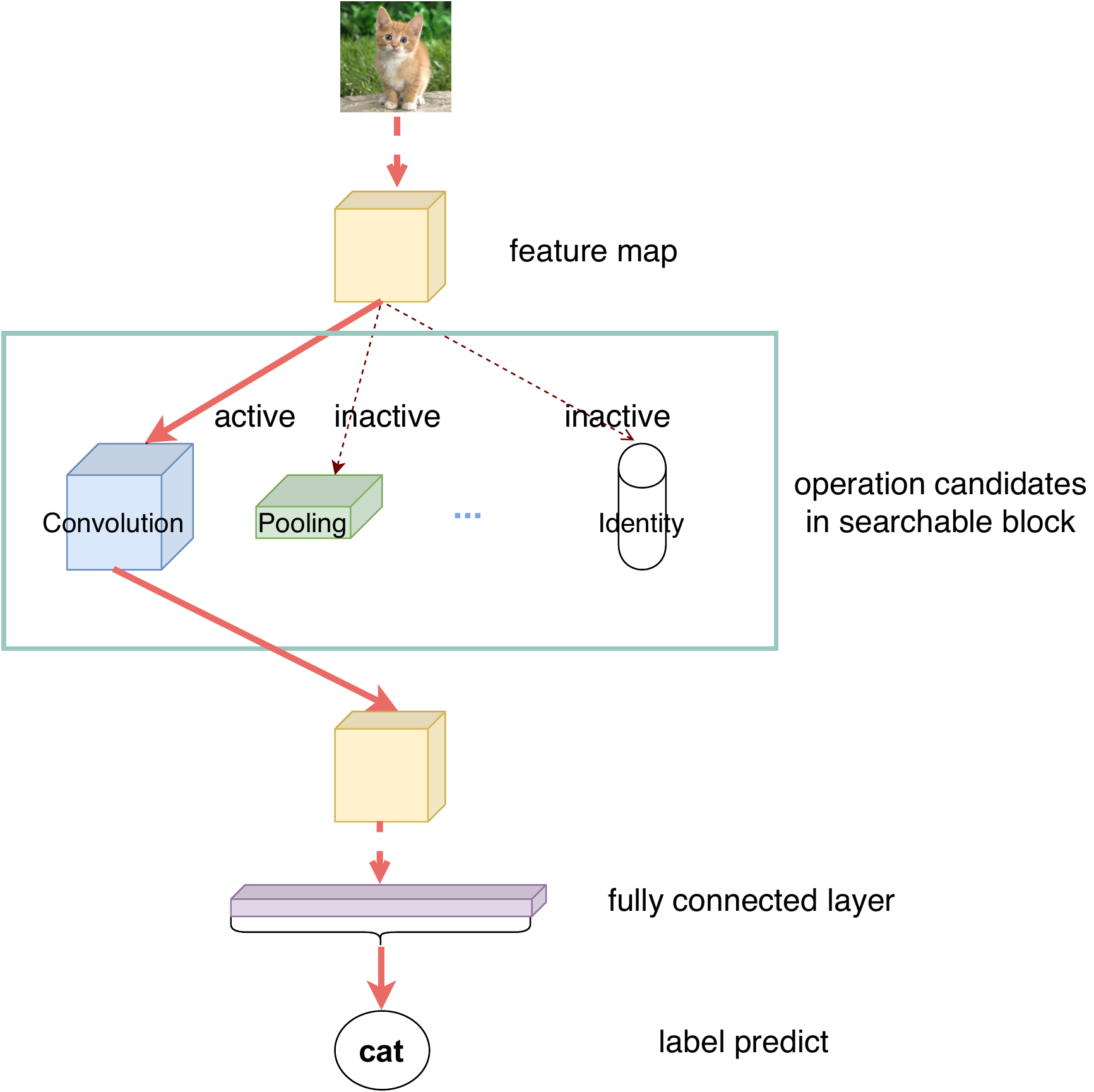}
    \caption{Illustration of ProxylessNAS.}
    \label{fig:proxyless-forward}
\end{figure}

In ProxylessNAS, an over-parameterized neural network is firstly constructed as shown in Fig. \ref{fig:proxyless-forward}, which is denoted by a directed acyclic graph (DAG) with $N$ nodes. Each node $x^{(i)}$ represents a latent representation (e.g., a feature map in convolutional neural networks), and each directed edge $e^{(i,j )}$ that connects node  $x^{(i)}$ and  $x^{(j)}$ defines the following operation:
\begin{align}
\label{plnas1}
x^{(j)} =  \sum_{n=1}^N b_n^{(i \rightarrow j)} o_n \left[ x^{(i)} \rightarrow x^{(j)} \right],
\end{align}
where $o_n\left[x^{(i)} \rightarrow x^{(j)}\right] \in \mathcal O$ denotes an operation candidate (e.g., convolution, pooling, identity, zero, ect.) that transforms $x^{(i)}$to $x^{(j)}$, and vector $\mathbf b^{(i \rightarrow j)} = [b_1^{(i \rightarrow j)}, \cdots, b_N^{(i \rightarrow j)}]$ is a binary gate taking values as follows:
\begin{align}
\label{plnas2}
\mathbf b^{(i \rightarrow j)} =     \begin{cases}
        [1, 0, \cdots, 0], & \text{with a probability $p_1^{(i \rightarrow j)}$}, \\
        ~~~~~~~~\cdots \\
        [0, 0, \cdots, 1], & \text{with a probability $p_N^{(i \rightarrow j)}$}. \\
    \end{cases}
\end{align}
{\color{black} From ~(\ref{plnas1}) and ~(\ref{plnas2}), it can be seen that rather than computing all the operations in the operation set $\mathcal O$, there is only one operation $o_n \left[ x^{(i)} \rightarrow x^{(j)}\right]$ that is utilized to transform each node $x^{(i)}$ to one of its neighbors $x^{(j)}$ with a probability $p_n^{(i \rightarrow j)}$ at each run-time, thus greatly saving the memory and computations in the training phase. In other words, only one path of activation is active in memory at run-time, resulting in $O(1)$ memory rather than $O(N)$ memory, which is much more efficient than other NAS counterparts, see, e.g., \cite{cai2018proxylessnas,cai2020once}. By this way, the memory requirement of training the over-parameterized network is  reduced to the same level of training a compact model, making ProxylessNAS a promising candidate for AIoT applications.} After learning these probabilities $\{p_n^{(i \rightarrow j)}\}_{n=1}^N$ from the training data and validation data, the operation $o_{n^*} \left[x^{(i)} \rightarrow x^{(j)}\right]$ with the largest probability $p_{n^*}^{(i \rightarrow j)} = \arg\max_{n=1,...,N} \{p_n^{(i \rightarrow j)}\}_{n=1}^N$  is chosen as the final operation for each directed edge $e^{(i \rightarrow  j )}$, while other operations are pruned out. 

\begin{figure*}[htbp]
    \centering
    \includegraphics[width=1\linewidth]{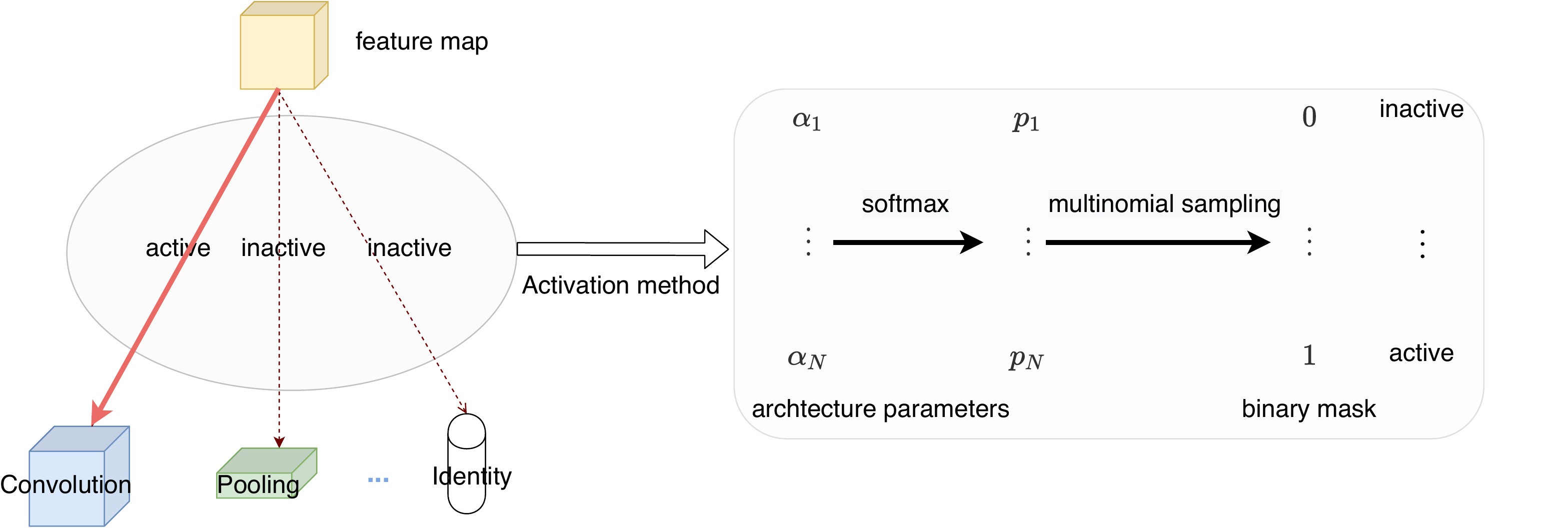}
    \caption{Operation activation in ProxylessNAS.}
    \label{fig:proxyless-binary-gate}
\end{figure*}
Despite the benefits of saving computational resources, ProxylessNAS faces a critical challenge: how to learn these probabilities $\{p_n^{(i \rightarrow j)}\}$ for each link $e^{(i \rightarrow  j )}$? Notice that these probabilities  $\{p_n^{(i \rightarrow j)}\}$ are constrained to lie  in a standard simplex $\{ \sum_{n=1}^N p_n^{(i \rightarrow j)}  = 1, p_n^{(i \rightarrow j)} \geq 0, \forall n \}$, making the direct optimization difficult. Instead, as Fig.~\ref{fig:proxyless-binary-gate} shows, inspired by the softmax function, a set of architecture parameters $\{\alpha_n^{(i \rightarrow j)}\}_{n=1}^N$, which are with simple nonnegative constraints $\alpha_n^{(i \rightarrow j)}  \geq 0, \forall n$, are introduced to equivalently represent each probability $p_n^{(i \rightarrow j)} $ as: 
\begin{align}
    p_n^{(i \rightarrow j)}  = \frac{\exp(\alpha_n^{(i \rightarrow j)} )}{\sum_{n^\prime=1}^N \exp(\alpha_{n^\prime}^{(i \rightarrow j)} )}.
\end{align}
This equivalent representation greatly facilitates the training algorithm design. {\color{black}
In particular, after taking the idea from BinaryConnect~\cite{courbariaux2015binaryconnect}, the gradient for each architecture parameter $\alpha_n^{(i \rightarrow j)} $ can be approximately computed by:}
\begin{equation}
\label{eq:binarybp}
    \frac{\partial \mathcal{L}}{\partial \alpha_n^{(i \rightarrow j)} }  \approx 
    \sum_{n^\prime = 1}^{N} \frac{\partial \mathcal{L}}{\partial b_{n^\prime}^{(i \rightarrow j)} } p_{n^\prime}^{(i \rightarrow j)}  (\sigma_{n, n^\prime}^{(i \rightarrow j)} - p_n^{(i \rightarrow j)}),
\end{equation} 
where $\mathcal{L}$ denotes the loss function, and $\sigma_{n, n^\prime}^{(i \rightarrow j)}$ takes its value as follows:
\begin{equation}
\label{eq:sigma_value}
    \sigma_{n, n^\prime}^{(i \rightarrow j)}=\begin{cases}
        $1$, & \text{with $n = n^\prime$}, \\
        $0$, & \text{with $n \neq n^\prime$}. \\
    \end{cases}
\end{equation}
In (\ref{eq:binarybp}), the gradient $\frac{\partial \mathcal{L}}{\partial b_{n^\prime}^{(i \rightarrow j)} }$ can be easily computed via back propagation since the binary gate $b_{n^\prime}^{(i \rightarrow j)} $ is involved in the computational graph.  However, to compute the gradients $\left \{\frac{\partial \mathcal{L}}{\partial b_{n^\prime}^{(i \rightarrow j)} } \right \}_{n^\prime = 1}^N$, all the $N$ operation candidates need to be activated, thus consuming  huge computational resources in the training phase.  {\color{black}To avoid this, in ProxylessNAS, an efficient and simple scheme was proposed. In particular, in each update step, two operation candidates are sampled according to the  probabilities $\{p_{n}^{(i \rightarrow j)}\}_{n=1}^N$ , while other operations are masked. Consequently, only two operation candidates would be taken into the computation, thus greatly enhancing the computational efficiency. Through this way,  regardless of the path number $N$, only two paths are involved in the update of architecture parameters, making the memory reduced to the same level of training a compact model.} In order to mitigate the effect from masking other operation candidates, rescaling all the architecture parameters is required at the end of each update step. Finally, the key steps of the training algorithm for ProxylessNAS is summarized in {\bf Algorithm \ref{alg:proxyless-nas}}~\cite{cai2018proxylessnas}.

In the framework of ProxylessNAS, the inference latency on hardware platforms can be naturally taken into account in order to increase the inference speed of the model. More specifically, for each link $e^{(i \rightarrow  j )}$ in the neural network, the expected latency can be given by:
\begin{align}
         \mathcal{L}_{\mathrm{latency}}^{(i \rightarrow j)} =\sum_{n=1}^N  p_n^{(i \rightarrow j)}  F\left(o_n \left[ x^{(i)} \rightarrow x^{(j)} \right] \right),
         \label{eq:layerlatency}
\end{align}
where $F (\cdot)$\footnote{ $F (\cdot)$  is obtained by referring to a latency lookup table that recording measured operations’ inference time on CPU, GPU, or mobile phone.} denotes the latency prediction model for operation candidates in the set $\mathcal O$. After taking the expected latency for all of the links in the neural network into account, the loss function for training takes the following form:
\begin{align}
         \mathcal L = \mathcal L_{\mathrm{CE}}  + \lambda_1 \sum_{\forall i \rightarrow j}   ||\mathbf w^{i \rightarrow j} ||_F^2 + \lambda_2 \sum_{\forall i \rightarrow j}  \mathcal{L}_{\mathrm{latency}}^{(i \rightarrow j)},
\end{align}
where $\mathcal{L}_{\mathrm{CE}}$ denotes classification cross-entropy loss, $\sum_{\forall i \rightarrow j}   ||\mathbf w^{i \rightarrow j} ||_F^2$ denotes the regularization for all of the neural network  parameters, and $\sum_{\forall i \rightarrow j}  \mathcal{L}_{\mathrm{latency}}^{(i \rightarrow j)} $ is the latency regularization for all the links. It has been shown that the incorporation of the latency regularization term still follows the training steps in {\bf Algorithm~\ref{alg:proxyless-nas}}, while with slight modifications on the gradient computation. More details could be found in {\bf Algorithm~\ref{alg:proxyless-nas}}.

\begin{algorithm}[t]\small
    \caption{ProxylessNAS~\cite{cai2018proxylessnas}}
 \label{alg:proxyless-nas}
    \begin{algorithmic} 
        \STATE {\bf Input:} An over-parameterized network $\{ x^{(i)}, e^{(i,j)} \}$.
    \STATE Initialize weight parameter $\mathbf w^{i\rightarrow j}, \forall i\rightarrow j$ and architecture parameter $\alpha_n^{i \rightarrow j}, \forall i \rightarrow j, \forall n.$ 
    \WHILE{not converged} 
    \STATE Sample each binary gate $\mathbf{b}^{(i \rightarrow j )}$ from $\alpha_n^{i \rightarrow j}, \forall n$.
    \STATE According to $\mathbf{b}^{(i \rightarrow j )},  \forall i \rightarrow j$, construct a SubNetwork with parameter $\mathbf w^{i\rightarrow j}, \forall i\rightarrow j$. 
    \STATE Update  $\mathbf w^{i\rightarrow j}, \forall i\rightarrow j$ using back propagation on training dataset.
    \STATE Update $\alpha_n^{i \rightarrow j}, \forall i \rightarrow j, \forall n$ using (4) on validation dataset.
    \ENDWHILE
    \end{algorithmic}
\end{algorithm}

\section{Federated  Direct Neural Architecture Search}
\label{sec:fdnas}

\subsection{Motivation and Problem Formulation}
In the last section, ProxylessNAS is introduced as an efficient framework to directly search neural network architectures on large-scale datasets without resorting to the knowledge from proxy tasks (e.g., training results on a small dataset). It is of great value for real-world applications since it is more efficient to use models directly searched from diverse daily data. However, the training algorithm of ProxylessNAS needs to collect all target data in advance, which inevitably limits its practical use because of the privacy concern in raw data transmission. In contrast, FedAvg trains a neural network without exchanging any data and can protect the data privacy of devices. However, the original form of FedAvg does not take the neural architecture search into account and requires cumbersome manual architecture design (or hyperparameter tuning) for the best performance.

{\color{black}To gain both the merits from ProxylessNAS and FedAvg while overcome their shortcomings, we develop a federated ProxylessNAS algorithm in this section. In particular, we follows the following problem formulation \cite{he2020fednas}:}
\begin{align}
    \min_{\pmb {\alpha}} \quad & \sum_{k=1}^K\mathcal{L}_{val}^{k}(\mathbf w^*(\pmb {\alpha}), \pmb {\alpha}) \label{eq:outerfednasFormulation}\\
    \text{s.t.} \quad &\mathbf w^*(\pmb {\alpha}) = \mathrm{argmin}_{\mathbf w} \enskip \sum_{k=1}^K\mathcal{L}_{train}^{k}(\mathbf w, \pmb {\alpha}), \label{eq:innerfednasFormulation}
\end{align}
where $\mathcal{L}_{val}^{k}(\cdot, \cdot)$ denotes the validation loss function of device $k$,   and $\mathcal{L}_{train}^{k}(\cdot, \cdot)$ denotes its training loss function. Vector $\pmb {\alpha}$ collects all architecture parameters $\{\alpha_n^{i \rightarrow j}, \forall i \rightarrow j, \forall n\}$ and  $\mathbf w$ collects all weights $\{\mathbf w^{i\rightarrow j}, \forall i\rightarrow j\}$. This nested bilevel optimization formulation is combined by those proposed in ProxylessNAS \cite{cai2018proxylessnas}, FedAvg \cite{pmlr-v54-mcmahan17a}. {\color{black}
FedNAS has some prior research effort in federated bilevel optimization~\cite{he2020fednas}.} More specifically, for each device, the goal is to search the optimal architecture $\pmb {\alpha}$ that gives the best performance on its local validation dataset, while with the optimal model weights $\mathbf w^*(\pmb {\alpha})$ learnt from its local training dataset. 

%

\subsection{Federated Algorithm Development}
\begin{figure*}[htbp]
    \centering
    \includegraphics[width=0.75\linewidth]{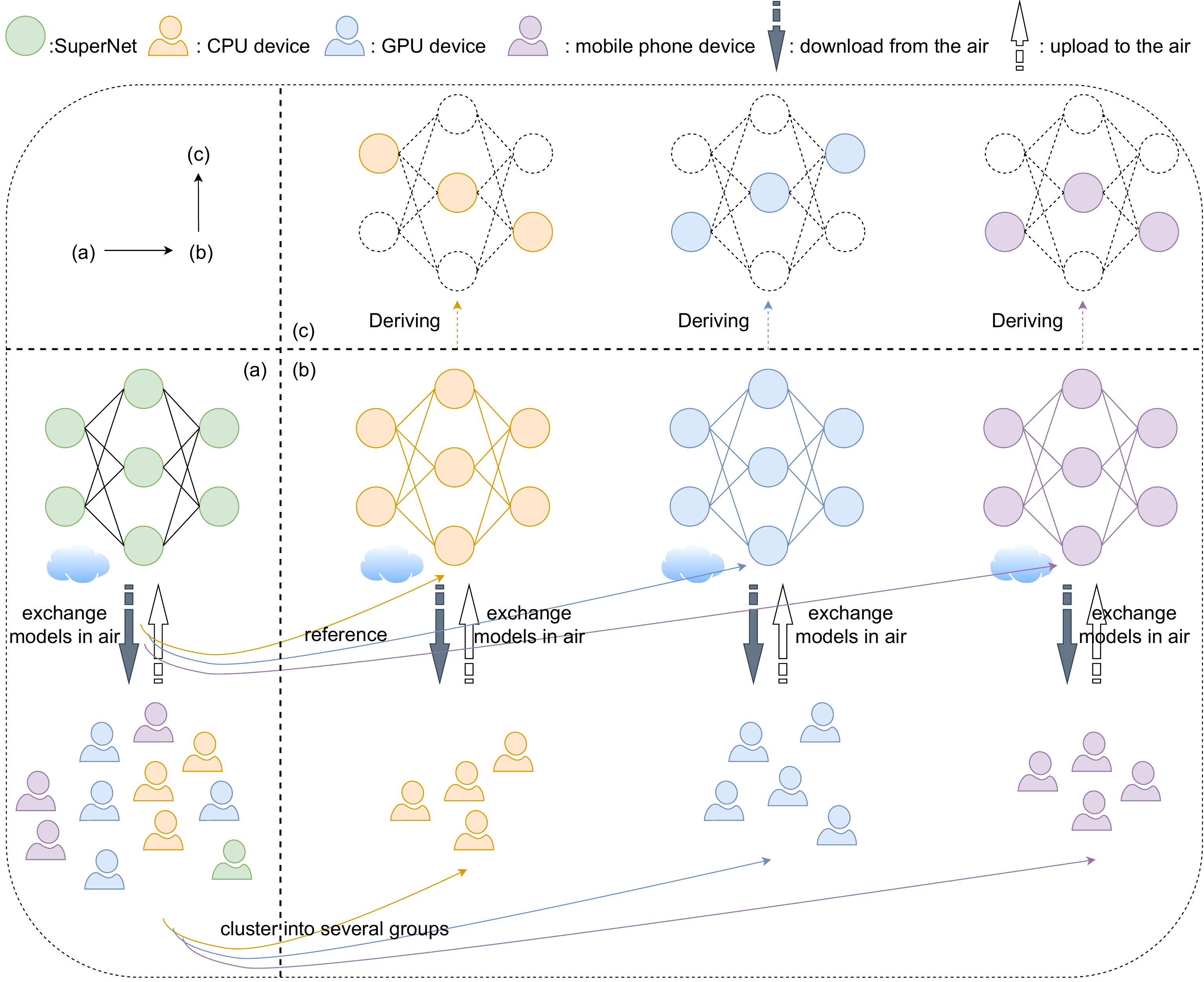}
    \caption{Illustration of (a) FDNAS, (b) CFDNAS and (c) deriving normal nets from CFDNAS SuperNets for AIoT devices.}
    \label{fig:fdnas}
\end{figure*}

To optimize $\pmb {\alpha}$ and $\mathbf w$ while protecting the privacy of device data, the FDNAS is described in this subsection. 

Following the set-up of federated learning, assume that there are $K$ devices in the set $S$ and one central server. In communication round $t+1$, each device downloads global parameters  $\pmb {\alpha}^g_t$ and $\mathbf w^g_t$ from the server and update these parameters using its local validation dataset and training dataset, respectively.  With global  parameters $\pmb {\alpha}^g_t$ and $\mathbf w^g_t$ as initial values, the updates follow the steps in {\bf Algorithm \ref{alg:fdnas}} {\color{black}\cite{he2020fednas}}, i.e.,  
\begin{align}
    \mathbf w_{t+1}^k, \pmb \alpha_{t+1}^k \leftarrow \text{ProxylessNAS}( \mathbf w_t^{g}, \pmb \alpha_t^{g}), \forall k \in S. \label{eq:w-alpha-from-single-client}
\end{align}
Then, each device sends its updated parameters $\{\mathbf w^{k}_{t+1},\pmb \alpha^{k}_{t+1} \}_{k \in S}$ to the central server for aggregation. Central server aggregates these parameters to update global parameters $\pmb {\alpha}^g_{t+1}$ and $\mathbf w^g_{t+1}$ {\color{black}\cite{he2020fednas}}:
\begin{align}
    \mathbf w_{t+1}^{g}, \pmb  \alpha_{t+1}^{g} \leftarrow \sum_{k=1}^\nc \frac{N_k}{N} \mathbf w_{t+1}^k, \sum_{k=1}^\nc \frac{N_k}{N} \pmb \alpha_{t+1}^k, \label{eq:sum-params-from-clients}
\end{align}
where $N_k$ is the size of local dataset in device $k$, and $N$ is the sum of all device's data size (i.e.,  $N = \sum_{k=1}^K N_k)$. 

After $T$ communication rounds, each device downloads the learnt global parameters $\pmb {\alpha}^g_t$ and $\mathbf w^g_t$ from the server, from which the optimized neural network architecture and model parameters can be obtained, as introduced in Section~\ref{sec:preliminary}. Furthermore, with the learnt neural network architecture, the weights $\mathbf w^k$ in each device could be further refined by the conventional FedAvg algorithm for better validation performance. The proposed algorithm is summarized in {\bf Algorithm \ref{alg:fdnas}} and illustrated in Fig. \ref{fig:fdnas}.

\begin{algorithm}[!h]
\caption{FDNAS for AIoT: Federated Direct Neural Architecture Search for AIoT}
\label{alg:fdnas}

\begin{algorithmic}
\SUB{Central server:}
   \STATE Initialize $\mathbf w_0^{g}$ and $\pmb \alpha_0^{g}$.
   \FOR{each communication round $t = 1, 2, \dots, T$}
     \FOR{each AIoT device $k \in S$ \textbf{in parallel}}
       \STATE $\mathbf w_{t+1}^k, \pmb \alpha_{t+1}^k \leftarrow \textbf{DeviceUpdate}(\mathbf w_t^{g}, \pmb \alpha_t^{g})$
     \ENDFOR
     \STATE $\mathbf w_{t+1}^{g}, \pmb \alpha_{t+1}^{g} \leftarrow \sum_{k=1}^\nc \frac{N_k}{N} \mathbf w_{t+1}^k, \sum_{k=1}^\nc \frac{N_k}{N} \pmb \alpha_{t+1}^k$
   \ENDFOR
   \STATE
   
 \SUB{DeviceUpdate($\mathbf w, \pmb \alpha$):} // \emph{On AIoT device platform.}
   \FOR{each local epoch $i$ from $1$ to $E$}
    \STATE update  $\mathbf w, \pmb \alpha$ by ProxylessNAS (Algorithm~\ref{alg:proxyless-nas})
  \ENDFOR
   \STATE return $\mathbf w, \pmb \alpha$ to server
\end{algorithmic}
\end{algorithm}

\subsection{Further Improvement and Insights}
\subsubsection{Clustering-aided model compression}
\begin{algorithm}[tb]
\caption{CFDNAS for AIoT: Cluster Federated Direct Neural Architecture Search for AIoT}
\label{alg:cfdnas}

\begin{algorithmic}
\SUB{Cluster server:}
   \STATE load $\mathbf w_0^{g}, \pmb \alpha_0^{g}$ by central server.
   \STATE $\{S_{1},\cdots,S_{P}\} \leftarrow$ (split $S_{all}$ into clusters by users' tag.)
   \FOR {cluster $S_{tag} \in \{S_{1},\cdots,S_{P}\}$ \textbf{in parallel}}
   \FOR{each round $t = 1, 2, \dots$}
     \FOR{each sampled device $k \in S_{tag}$ \textbf{in parallel}}
       \STATE $\mathbf w_{t+1}^k, \pmb \alpha_{t+1}^k \leftarrow \text{DeviceUpdate}_k(\mathbf w_t^{g}, \pmb \alpha_t^{g})$ 
     \ENDFOR
     \STATE $\mathbf w_{t+1}^{g}, \pmb \alpha_{t+1}^{g} \leftarrow \sum_{k=1}^\nc \frac{n_k}{n_{all}} \mathbf w_{t+1}^k, \sum_{k=1}^\nc \frac{n_k}{n_{all}} \pmb \alpha_{t+1}^k$
   \ENDFOR
   \ENDFOR
   \STATE

\SUB{DeviceUpdate($\mathbf w, \pmb \alpha$):} // \emph{On device platform.}
  \FOR{each local epoch $i$ from $1$ to $E$}
    \STATE update  $\mathbf w, \pmb \alpha$ by ProxylessNAS
 \ENDFOR
 \STATE return $\mathbf w, \pmb \alpha$ to server
\end{algorithmic}
\end{algorithm}

{\color{black}
At the end of the proposed algorithm (i.e., {\bf Algorithm \ref{alg:fdnas}}), the neural network architecture is learnt from the datasets of all the devices in a federated way like a meta-train phase~\cite{finn2017model, nichol2018first}} Although it enables knowledge transfer, accounting for every piece of information might result in structure redundancy for a particular group of devices. More specifically, assume that ten devices collaboratively train a model for image classification. Some devices are with images about \textit{birds, cats, and deer} and thus can form a ``animal" group, while other devices in the ``transportation" group are with images about \textit{airplanes, cars, ships, and trucks}. Although these images all contribute to the neural network structure learning in  {\bf Algorithm \ref{alg:fdnas}}, the operations tailored for the ``animal" group might not help the knowledge extraction from the ``transportation" group significantly. Therefore, an {\bf immediate idea} is: \emph{
{\color{black}
could we further refine the model architecture in an efficient way by utilizing the clustering information of devices?}}

To achieve this, we further propose a clustering-aided refinement scheme into the proposed FDNAS. In particular, after $\pmb \alpha^{g}$ converges, each device could send a tag about its data to the server, e.g.,  \textit{animal, transportation}, which are not sensitive about its privacy. Then the server gets the clustering information about the devices, based on whether the devices are with similar data distributions, and then divides the device set $S$ into several groups:
\begin{align}
    S \rightarrow \{S_{1},\cdots,S_{P}\}.
    \label{eq:s-to-si}
\end{align}
{\color{black} Note that these tags are {\it not} the labels of training data. Particularly, each tag only provides a very general and abstract summarization of data (such as ``animal'' and ``transportation''), but does not convey the detail label information of the training data (such as ``cat'' and ``airplane''). Therefore, it can be used in the scenarios  that these general tags are not treated as the private information but the terms that can be shared.  Finally, the devices in the same group could further refine their SuperNet by re-executing the proposed FDNAS algorithm like the meta-test phase.}
The proposed clustering-aided refinement scheme labeled as \textbf{CFDNAS} is summarized in {\bf Algorithm \ref{alg:cfdnas}}.  

\subsubsection{Latency-aware model compression}
Devices might deploy models on different devices, such as mobile phones, CPUs, and GPUs. Previous studies show that taking latency-ware loss into account could guide the NAS to the most efficient one in terms of the inference speed. 
For example, ProxlessNAS taking the latency-aware loss into account can skip the computationally expensive operations in each layer. 
The proposed FDNAS and CFDNAS could naturally integrate the hardware latency information into its search scheme. In particular, we could let each device send their hardware information such as \textit{GPU, CPU} configuration to the server. Then, the server divides these devices into several groups according to their hardware type. For the devices in the same group, a latency-ware loss term is added to the training loss. By taking this latency-ware loss term into the training process, the proposed algorithm will drive the SuperNet to a compact one that gives the fastest inference speed for the particular hardware platform.

{\color{black} {\it Remark 1:} Overall, we view the prior FDNAS (see Algorithm 3) as meta-train phase in meta learning, which was trained on all devices. After finishing the FDNAS SuperNet training, we view the CFDNAS as a meta-test phase. Particularly, in CFDNAS, SuperNet is initialized based on the learning result of FDNAS. Then, devices within each cluster utilize the SuperNet to quickly search a tailored neural network after only a few rounds of updates. This is how the idea of meta learning comes into the design of the proposed NAS scheme.  

{\it Remark 2:} The FDNAS SuperNet is trained across all devices but not limited to a single cluster of devices. This FDNAS SuperNet has two contributions: first, it helps CFDNAS SuperNets converge fast, thus costing little extra training time; second, it helps normal nets derived from CFDNAS SuperNets achieve good trade-offs between accuracy and inference time compared with the ones directly derived from FDNAS, see more results in Table VI.}

\begin{table*}[!htbp]
\centering
\caption{\textbf{CIFAR-10 performance.} $^*$: The federated averaged model's accuracy. $\dag$: mean local accuracy, i.e., each device completes training local epochs after downloading the server model, and then averages the local accuracies obtained by testing inference on the local test data. $^{\circ\circ\circ}$: 15 local epochs for each device. $^{\circ\circ}$: 10 local epochs for each device. $^{\circ}$: 5 local epochs for each device.}
\begin{tabular}{lcccc}
\toprule
\textbf{\multirow{2}{*}{Architecture}} &\textbf{\multirow{2}{*}{Test Acc. (\%)}}   & \textbf{Params.} & \textbf{Search Cost} & \textbf{\multirow{2}{*}{Method}} \\
\cmidrule(lr){3-4} &
 & \textbf{(M)} & \textbf{(GPU hours)} &\\
\hline
DenseNet-BC~\cite{densenet}                       & 94.81  & 15.3 & -    & manual \\
MobileNetV2~\cite{sandler2018mobilenetv2}                                   & 96.05  & 2.5   & -  & manual \\
\hline
NASNet-A~\cite{zoph2018learning}                 & 97.35      & 3.3  & 43.2K & RL      \\
AmoebaNet-A~\cite{amoebanet}           & 96.66      & 3.2  & 75.6K & evolution \\
AmoebaNet-B~\cite{amoebanet}           & 97.45   & 2.8  & 75.6K & evolution \\
Hireachical Evolution~\cite{liu2018hierarchical}          & 96.25   & 15.7 & 7.2K  & evolution \\
PNAS~\cite{liu2018progressive}                            & 96.59    & 3.2  & 5.4K  & SMBO \\
ENAS~\cite{pham2018efficient}                    & 97.11     & 4.6  & 12  & RL \\
\hline
DARTS~\cite{liu2018darts}          & 97.24 & 3.3 & 72 & gradient \\
SNAS~\cite{xie2018snas}        & 97.02 & 2.9  & 36  & gradient \\
P-DARTS C10~\cite{chen2019progressive}                                & 97.50 & 3.4  & 7.2 & gradient \\
P-DARTS C100~\cite{chen2019progressive}                                & 97.38 & 3.6  & 7.2 & gradient \\
PC-DARTS~\cite{xu2020pcdarts} & 97.39 &3.6 &2.5 &  gradient\\
GDAS~\cite{dong2019searching}                                   & 97.07 & 3.4  & 5 & gradient \\
\hline
{\color{black}
\textbf{FDNAS (ours)$^{\circ\circ\circ}$} }                              & \textbf{72.79$^*$/95.88$\dag$}  & \textbf{3.5}  & \textbf{135} & \textbf{gradient \& federated} \\
{\color{black}
\textbf{FDNAS (ours)$^{\circ\circ}$} }                              & \textbf{74.12$^*$/96.52$\dag$}  & \textbf{3.5}  & \textbf{85} & \textbf{gradient \& federated} \\
{\color{black}
\textbf{FDNAS (ours)$^{\circ}$} }                                 & \textbf{78.75$^*$/97.25$\dag$}  & \textbf{3.4}  & \textbf{59} & \textbf{gradient \& federated} \\

\bottomrule
\end{tabular}
\label{tab_ev_cifar}
\end{table*}

\section{Experiments}
\label{sec:exp}
\subsection{Implementation Details}
We use PyTorch \cite{paszke2019pytorch} to implement FDNAS and CFDNAS.
We searched on the CIFAR-10 dataset~\cite{cifar10} and then trained normal nets from scratch. CIFAR-10 has 50K training images, 10K test images, and ten classes of labels. To simulate an AIoT scenario, we set up ten devices. The first three classes of images are randomly assigned to the first three devices. Then, the middle three classes of images are randomly assigned to the middle three devices, and the last four classes of images are randomly assigned to the last four devices. 
Each device has 4500 images as a training set for learning $\mathbf w$ and 500 images as a validation set for learning $\pmb \alpha$.
Besides, to test the transferability of FDNAS after searching on CIFAR-10, we train FDNAS normal net on ImageNet dataset~\cite{5206848}.
{\color{black}
The image classification dataset LEAF FEMNIST has 3,550 devices and 80,5263 images~\cite{caldas2018leaf}. There are 226.83 samples per device (mean) in the context of FDNAS.
The image classification dataset FedML is built on CIFAR-10 that 50K images are used as training samples and 10K images are used as testing samples~\cite{he2020fedml}. In FedML, we define 16 devices and apply Latent Dirichlet Allocation as the partition method to assign non-IID data points for devices.}

\subsection{Image Classification on CIFAR-10}
\subsubsection{Training Settings}
We use a total batch size of 256 and set the initial learning rate to $0.05$. Then, we use the cosine rule as a decay strategy for the learning rate. When training SuperNet, we use Adam to optimize $\pmb \alpha$ and momentum SGD to optimize $\mathbf w$.\footnote{\color{black} The choice of optimizer for  weight parameters and architecture parameters follows the convention in previous studies~\cite{liu2018darts, cai2018proxylessnas, xu2020pcdarts,yang2019evaluation}. That is, SGD is usually adopted to optimize weight parameters, and Adam is widely used to optimize architecture parameters. } The weight decay of $\mathbf w$ is $3\times10^{-4}$, while we do not use weight decay for $\pmb \alpha$.
SuperNet has 19 searchable layers and is composed in the same way as ProxylessNAS (each layer consists of MBConv blocks)~\cite{cai2018proxylessnas}.
We train FDNAS SuperNet on ten online devices for 125 rounds. Meanwhile, the local epoch of each device is 5 (default) and 10. 
After that, CFDNAS clusters  device-0, 1, and 2 into the GPU group to train CFDNAS-G SuperNet, and  device-3, 4, and 5 into the CPU group to train CFDNAS-C SuperNet. CFDNAS SuperNets are all searched separately for 25 rounds.
After training the SuperNet, we derive the normal net from the SuperNet and then run 250 rounds of training from scratch on devices via FedAvg. 
{\color{black} For the workstation-level desktop,
GPU latency is measured on TITAN Xp GPUs with a batch size of 128 to avoid severe GPU underutilization caused by small batches, and CPU latency is measured on two 2.20GHz Intel(R) Xeon(R) E5-2650 v4 server CPUs with a batch size of 128.
For light mobile device, that is, ultrabook-level laptop, GPU latency is measured on GTX MX150 and CPU latency is measured on the Intel i7-8550U CPU.
On the mobile phone, we choose Samsung Galaxy S20 as the hardware platform and set the batch size of 1 following previous works~\cite{cai2018proxylessnas, cai2020once}.
}

\subsubsection{Results}
Since FDNAS is based on FedAvg and protects the data privacy of the device, we put both the federated averaged accuracy of models and local mean accuracy of all devices in Table~\ref{tab_ev_cifar}. 
{\color{black} The FDNAS with local epoch number 5 makes the accuracy of searched normal nets slightly higher than the FDNAS with local epoch number 10 and 15. Thus, we set local epoch number's default value as 5.}
As Table~\ref{tab_ev_cifar} demonstrated,  
FDNAS normal net achieves $78.75\%$ federated averaged accuracy and $97.25\%$ mean local accuracy. The federated learning framework is naturally suited to distributed training, and it only costs 59 GPU hours. 
FDNAS outperforms both evolution-based NAS and gradient-based DARTS in search time cost. Also, the local accuracy of FDNAS is higher than their accuracies.
Besides, we use MobileNetV2 as a pre-defined, hand-crafted model trained under FedAvg with the same MBconv-block setting to compare with FDNAS. In Table~\ref{ablation1},  FDNAS outperforms MobileNetV2 in both federated averaged accuracy and mean local accuracy. See Subsection~\ref{alation-sec} for more ablation studies and the performance of CFDNAS.

{\color{black}
\subsection{Image Classification on LEAF}
\subsubsection{Training Settings}
We apply a batch size of 32. The initial learning rate is $0.005$ and decays under cosine rule. When training SuperNet, we use Adam to optimize $\pmb \alpha$ and momentum SGD to optimize $\mathbf w$. The weight decay of $\mathbf w$ is $4\times10^{-5}$, while we do not use weight decay for $\pmb \alpha$. 
\subsubsection{Results}
Since the setting of device number and data distribution is based on LEAF FEMNIST, we compare FDNAS with default models in LEAF. The depth of search space in SuperNet is similar to the depth of default models in LEAF for a fair comparison. By FDNAS, the model directly searched on LEAF FEMNIST obtains better trade-offs among the amount of parameters, FLOPs and accuracy,  compared with LEAF-CNN and human-craft MobileNetV2, as seen in Table II.

\subsubsection{Robustness against Stragglers:} To analyze the impact on the proposed method in the presence of the stragglers, we  view the  stragglers as offline clients/devices in each round, and evaluate their influence on training SuperNet and normal net. Particularly, using LEAF dataset, when stragglers exist (35 devices (of total 3,550 devices) are randomly sampled to be online in each global round),  the performance of  FDNAS and those of  other human-craft neural architectures are presented in Table II. It can be seen that the proposed method is robust against the existence of stragglers.

\begin{table*}[!htbp]
\centering
{\color{black}
\caption{\textbf{LEAF FEMNIST performance.} $^*$: The federated averaged model's accuracy.} }
\begin{tabular}{lcccc}
\toprule
\textbf{\multirow{2}{*}{Architecture}} &\textbf{\multirow{2}{*}{Test Acc. (\%)}}   & \textbf{Params.} & \textbf{$\times+$} & \textbf{\multirow{2}{*}{Method}} \\
\cmidrule(lr){3-4} &
 & \textbf{(M)} & \textbf{(M)} &\\
\hline
{\color{black} LEAF-CNN ~\cite{caldas2018leaf} }  & 74.55$^*$  & 6.50 & 13.00    & manual \\
{\color{black} MobileNetV2~\cite{sandler2018mobilenetv2}  }                              & 73.70$^*$  & 3.30   & 8.13  & manual \\
\hline
{\color{black} \textbf{FDNAS (ours)} }                                 & \textbf{75.84$^*$}  & \textbf{3.43}  & \textbf{6.46} & \textbf{gradient \& federated} \\
\bottomrule
\end{tabular}
\label{tab_ev_leaf}
\end{table*}

\subsection{Image Classification on FedML}
\subsubsection{Training Settings}
We apply a batch size of 128. The initial learning rate is $0.025$ and decays under cosine rule. When training SuperNet, we use Adam to optimize $\pmb \alpha$ and momentum SGD to optimize $\mathbf w$. The weight decay of $\mathbf w$ is $3\times10^{-4}$, while we do not use weight decay for $\pmb \alpha$.
\subsubsection{Results}
Since the setting of device number and data distribution is based on FedML CIFAR-10, we compare FDNAS with FedNAS in FedML. The local epoch (15), global round (50) and SGD optimizer are  similar to the default hyper-parameters in FedML for a fair training comparison. The model directly searched on FedML CIFAR-10 by FDNAS gets better trade-offs between computational cost (FLOPs) and accuracy,  compared with FedNAS, human-craft MobileNetV2 and DenseNet, as seen in Table III.
\begin{table*}[!htbp]
\centering
\caption{
{\color{black}
\textbf{FedML CIFAR-10 performance.} $^*$: The federated averaged model's accuracy.}
}
\begin{tabular}{lcccc}
\toprule
\textbf{\multirow{2}{*}{Architecture}} &\textbf{\multirow{2}{*}{Test Acc. (\%)}}   & \textbf{Params.} & \textbf{$\times+$} & \textbf{\multirow{2}{*}{Method}} \\
\cmidrule(lr){3-4} &
 & \textbf{(M)} & \textbf{(M)} &\\
\hline
{\color{black} DenseNet~\cite{densenet} }                                 & 77.7$^*$  & -   & -  & manual \\
{\color{black} MobileNetV2~\cite{densenet}}                                   & 74.6$^*$  & 2.55   & 289.6  & manual \\
{\color{black}FedNAS ~\cite{caldas2018leaf} }  & 81.3$^*$  & 2.40 & 371.3    & gradient \& federated \\
\hline
{\color{black} \textbf{FDNAS (ours)} } & \textbf{81.8$^*$}  & \textbf{3.20}  & \textbf{330.2} & \textbf{gradient \& federated} \\
\bottomrule
\end{tabular}
\label{tab_ev_leaf}
\end{table*}

}
\subsection{Image Classification on ImageNet}
\begin{table*} [!t]
\centering
\caption{\textbf{ImageNet performance.} $\times+$ denotes the number of multiply-add operations (FLOPs).}
\begin{tabular}{lcccccc}
\toprule
\textbf{\multirow{2}{*}{Architecture}} & \multicolumn{2}{c}{\textbf{Test Acc. (\%)}} & \textbf{Params.} & $\times+$ & \textbf{Search Cost} & \textbf{\multirow{2}{*}{Search Method}} \\
\cmidrule(lr){2-3}
& \textbf{top-1} & \textbf{top-5} & \textbf{(M)} & \textbf{(M)} & \textbf{(GPU hours)} &\\
\hline
MobileNet~\cite{howard2017mobilenets}         & 70.6 & 89.5 & 4.2 & 569  & -    & manual \\
ShuffleNet 2$\times$ (v2)~\cite{ma2018shufflenet}    & 74.9 & - & $\sim$5  & 591  & -    & manual \\
MobileNetV2~\cite{sandler2018mobilenetv2}         & 72.0 & 90.4 & 3.4 & 300  & -    & manual \\

\hline
NASNet-A~\cite{zoph2018learning}              & 74.0 & 91.6  & 5.3 & 564  & 43.2K & RL \\
NASNet-B~\cite{zoph2018learning}              & 72.8 & 91.3  & 5.3 & 488  & 43.2K & RL \\
NASNet-C~\cite{zoph2018learning}              & 72.5 & 91.0  & 4.9 & 558  & 43.2K & RL \\
AmoebaNet-A~\cite{amoebanet}        & 74.5 & 92.0  & 5.1 & 555  & 75.6K & evolution \\
AmoebaNet-B~\cite{amoebanet}        & 74.0 & 91.5  & 5.3 & 555  & 75.6K & evolution \\
AmoebaNet-C~\cite{amoebanet}        & 75.7 & 92.4  & 6.4 & 570  & 75.6K & evolution \\
PNAS~\cite{liu2018progressive}                & 74.2 & 91.9  & 5.1 & 588  & 5.4K  & SMBO \\
MnasNet~\cite{tan2019mnasnet}              & 74.8 & 92.0  & 4.4 & 388  & -    & RL \\
\hline
DARTS~\cite{liu2018darts}      & 73.3 & 91.3  & 4.7 & 574  & 96    & gradient \\
SNAS~\cite{xie2018snas}     & 72.7 & 90.8  & 4.3 & 522  & 36  & gradient \\
ProxylessNAS~\cite{cai2018proxylessnas}       & 75.1 & 92.5  & 7.1 & 465  & 200  & gradient \\
P-DARTS-C10~\cite{chen2019progressive}                 & 75.6 & 92.6  & 4.9 & 557  & 7.2  & gradient \\
P-DARTS-C100~\cite{chen2019progressive}             & 75.3 & 92.5  & 5.1 & 577  & 7.2  & gradient \\
GDAS~\cite{dong2019searching}           & 74.0 & 91.5 & 5.3 & 581  & 5  & gradient \\\hline

{\color{black}
\textbf{FDNAS (mean)$^{\circ}$} }            & \textbf{75.3} & \textbf{92.9}  & \textbf{5.1} & \textbf{388}  & \textbf{59}  & \textbf{gradient~ \& federated} \\

\bottomrule
\end{tabular}
\label{ev_imagenet}
\end{table*}
\subsubsection{Training Settings}
We moved the FDNAS normal net to ImageNet for training to evaluate the generalization performance on larger image classification tasks. Following the general mobile setting~\cite{liu2018darts}, we set the input image size to $224\times224$. We set the FDNAS normal net's Layer 1, 3, 6, 8, and 16 as the downsampling layers. The model's FLOPs are constrained to below 600M.
We use an SGD optimizer with a momentum of 0.9. The initial learning rate is 0.4 and decays to 0 by the cosine decay rule. Then the weight decay is kept consistent at $4\times10^{-5}$. The dropout rate is 0.2.

\subsubsection{Results}
{\color{black}
As shown in Table~\ref{ev_imagenet}, we achieve SOTA performance compared to other methods. We use 3 different random seeds to initialize the FDNAS normal net's weight parameters and obtain  $75.34\%$, $75.32\%$ and $72.28\%$.
}
FDNAS mean test accuracy is $75.3\%$, which is higher than GDAS, ProxylessNAS, SNAS, and AmoebaNet. Besides, FDNAS normal net FLOPs are 388M, which also smaller than other methods. The search cost is only 59 GPU hours, which is smaller than ProxylessNAS and makes sense in real-world deployments.
Also, for comparison, we compare FDNAS to MobileNetV2 since they are both composed of MBconv blocks. FDNAS normal net achieves $75.3\%$ accuracy, which outperforms MobileNetV2 by $3.3\%$. Moreover, the FLOPs of MobileNetV2(1.4) is 585M, which is much more dense than 388M FLOPs of FDNAS. However, the accuracy of FDNAS still outperforms MobileNetV2(1.4) by $0.6\%$.
Summing up the above analysis, the model searched by FDNAS in a privacy-preserving manner is highly transferable, and it achieves an excellent trade-off between accuracy and FLOPs. These show that learning the neural architecture from the data can mitigate the bias caused by human effort and thus attain higher efficiency.

{\color{black} Note that for the test accuracy, since the proposed FDNAS/CFDNAS scheme has the similar search space with other benchmarking schemes (including ProxylessNAS and DARTS), it is expected that their test accuracies are similar, see discussions in \cite{yang2019evaluation,xie2019exploring}. Even with similar test accuracies, the proposed scheme allows tailored NAS for each device in a federated manner, thus achieving much more efficient normal nets in terms of memory cost and inference speed,  while at the same time protecting the users' privacy.}

\subsection{Ablation Study}
\label{alation-sec}
\begin{table*}[!t]
\centering
\caption{\textbf{Comparison among MobileNetV2, FDNAS and CFDNAS on CIFAR-10}: $\dag$ and $^*$ is explained in Table~\ref{tab_ev_cifar}. 
{\color{black}
$^\P$ denotes inference latency measured on the workstation and $^\S$ denotes inference latency measured on the ultrabook-level laptop. The latency on the phone is measured on Samsung Galaxy S20.}
}
\begin{tabular}{lcccccccccc}
\toprule
\textbf{\multirow{2}{*}{Architecture}} & {\color{black}\textbf{Test Set Partition}} & \multicolumn{2}{c}{\textbf{Lat. (ms}) $^\P$} & \multicolumn{2}{c}{\textbf{
{\color{black}
Lat. (ms) $^\S$ }
}
} &\textbf{{\color{black}Lat. (ms)}}  & \textbf{Params.}  & \textbf{$\times+$} & \textbf{Search Cost} & \textbf{\multirow{2}{*}{{\color{black}Test Acc.(\%)}}} \\ 
\cmidrule(lr){3-4}\cmidrule(lr){5-6}\cmidrule(lr){7-7}
& {\color{black}\textbf{of Device ID}} & \textbf{GPU} & \textbf{CPU} & \textbf{{\color{black}GPU}} & \textbf{{\color{black}CPU}} & \textbf{\color{black}Phone} & \textbf{(M)} & \textbf{(M)} & \textbf{(GPU hours)} &\\ \hline
    & 0,...,9   &    &    &   &   &  &  &                                               &&  68.45$^*$/ 96.51$\dag$\\ 
 MobileNetV2   & {\color{black}0,1,2}   & 52.31  &  890.69 & 249.26  &  2400.32 &110.20  &2.5 &296.5  & -      &  {\color{black}67.57$^*$/ 96.22$\dag$}\\ 
    & {\color{black}3,4,5}   &    &    &   &    & &  &                                           &      &  {\color{black}65.23$^*$/ 91.27$\dag$}\\ \hline
    
        & 0,...,9   &    &   &    &    &    &  &                                  &      &  78.75$^*$/ 97.25$\dag$\\
 FDNAS  & {\color{black}0,1,2} & 52.78  &  600.17& 233.45  &  1862.21 & 105.90 &  3.4 &346.6  &  59.00          &  {\color{black}73.50$^*$/ 98.79$\dag$}\\
        & {\color{black}3,4,5}   &    &   &    &   &    &   &          &                                &  {\color{black}71.90$^*$/ 93.01$\dag$}\\ \hline
        
 CFDNAS-G  & 0,1,2 & 40.33 & 463.86 & 212.30 & 1468.18 & 79.50  &  3.3  &318.4 &  3.53           &  73.60$^*$/ 98.93$\dag$\\\hline
 CFDNAS-C    & 3,4,5   & 31.00 & 186.52 & 143.02 & 863.00  &58.70 &  2.0 &169.3 &  3.46        &  71.29$^*$/ 93.01$\dag$\\
\bottomrule
\end{tabular}

\label{ablation1}
\end{table*}
\begin{table*}[!t]
\centering
\caption{\textbf{Enhancement by CFDNAS}. ``naive-CFDNAS'' means that SuperNet for CFDNAS is not inherited from FDNAS, but is searched directly in the cluster group from scratch. ``Personalized-FDNAS'' denotes that FDNAS SuperNet can directly derive out a normal net on each single device from personalized FDNAS SuperNets, which are fine-tuned on each single device local dataset based on FDNAS.}
\begin{tabular}{lccccccc}
\toprule
\textbf{\multirow{2}{*}{Architecture}} &\multirow{2}{*}{\textbf{Device ID}} & \textbf{Params.}  & \textbf{$\times+$} &\textbf{Search Cost}  &{\textbf{Device Local}} &\textbf{Mean Local} \\ 
&       &                     \textbf{(M)} & \textbf{(M)} &\textbf{(GPU hours)}&\textbf{Acc.(\%)}   & \textbf{Acc.(\%)} &\\ \hline

FDNAS   &0,1,2    &  3.38     & 346.64   &59 & 98.14/99.35/98.90    & 98.79   \\ 
{\color{black}Personalized-FDNAS}   &{\color{black}0,1,2}    &  {\color{black}3.55}     & {\color{black}354.32}   &{\color{black}60} & {\color{black}97.73/98.56/98.33}    & {\color{black}98.20}   \\ 
naive-CFDNAS-G   
   &0,1,2    & 3.70      & 356.83   &18 &97.56/98.11/97.36  &97.67  \\ 
\textbf{CFDNAS-G}   
   &0,1,2    & \textbf{3.33}      & \textbf{318.44}   &\textbf{3.53} &\textbf{98.39/99.02/99.38}  &\textbf{98.93}  \\ \hline
FDNAS    &3,4,5    &3.38 &346.64 &59 & 94.20/91.81/93.04    & 93.01   \\ 
{\color{black}Personalized-FDNAS}    &{\color{black}3,4,5}    &{\color{black}3.07} &{\color{black}300.04} &{\color{black}60} & {\color{black}93.00/91.57/92.51}    &{\color{black} 92.36}   \\ 
naive-CFDNAS-C   &3,4,5    & 1.92     & 187.89   &18 &88.61/89.88/90.25  &89.58  \\ 
\textbf{CFDNAS-C}  &3,4,5    & \textbf{2.03}      & \textbf{169.35}   &\textbf{3.46} &\textbf{93.21/92.73/93.12}  &\textbf{93.01}  \\
\bottomrule

\end{tabular}
\label{ablation3}
\end{table*}
\begin{table*}[!t]
\centering
\caption{\textbf{Compared with conventional DNAS.} $\dag$ and $^*$ is explained in Table~\ref{tab_ev_cifar}. The DNAS search algorithm is ProxylessNAS like FDNAS, but uses well-collected data rather than federated learning.}
\begin{tabular}{lccccc}
\toprule
\textbf{\multirow{2}{*}{Architecture}} & \textbf{Params.}  & \textbf{$\times+$} &\textbf{Search Cost}  & \textbf{Test Acc.} \\ 
&   \textbf{(M)} & \textbf{(M)} &\textbf{(GPU hours)}& \textbf{(\%)}  \\ \hline
DNAS(on device-0)       & 2.54      & 264.35   &6.4 & 95.83        \\ 
DNAS(on device-1)       & 3.50      & 330.58   &6.5 & 97.04        \\ 
DNAS(on device-2)       & 2.85      & 269.76   &8.4 & 97.55        \\ 
DNAS(on device-3)       & 2.77      & 258.90   &9.9 & 87.33        \\ 
DNAS(on device-4)       & 2.28      & 233.99   &6.7 & 88.38        \\ 
DNAS(on device-5)       & 3.27      & 299.63   &7.5 & 87.53        \\ 
DNAS(on device-6)       & 2.78      & 297.03   &7.7 & 97.83        \\ 
DNAS(on device-7)       & 3.25      & 336.20   &7.1 & 96.83        \\ 
DNAS(on device-8)       & 2.68      & 263.89   &8.8 & 97.38        \\ 
DNAS(on device-9)       & 2.80      & 286.46   &7.8 & 97.47        \\ 
mean           & 2.87      & 284.08   &7.7 & 94.31        \\ \hline

DNAS(on well-collected CIFAR-10)           & 5.02      & 500.71   &24.5 & 96.71        \\ \hline
{\color{black}FDNAS(on device-0)}          &3.38      &346.64     &59.0  &{\color{black}    98.14$\dag$}\\
{\color{black}FDNAS(on device-1)}          &3.38      &346.64     &59.0 &{\color{black}    99.35$\dag$}\\
{\color{black}FDNAS(on device-2)}          &3.38      &346.64     &59.0 &{\color{black}    98.90$\dag$}\\
{FDNAS(mean on device-0, 1, 2)}        &3.38      &346.64     &59.0  &{\color{black}      73.50$^*$/98.79$\dag$}\\ 
{\color{black}FDNAS(on device-3)}          &3.38      &346.64     &59.0  &{\color{black}    94.20$\dag$}\\
{\color{black}FDNAS(on device-4)}          &3.38      &346.64     &59.0   &{\color{black}    91.81$\dag$}\\
{\color{black}FDNAS(on device-5)}          &3.38      &346.64     &59.0   &{\color{black}    93.04$\dag$}\\
{FDNAS(mean on device-3, 4, 5)}        &3.38      &346.64     &59.0   &{\color{black}    71.90$^*$/93.01$\dag$}\\ 
{\color{black}FDNAS(on device-6)}          &3.38      &346.64     &59.0   &{\color{black}    98.97$\dag$}\\
{\color{black}FDNAS(on device-7)}          &3.38      &346.64     &59.0  &{\color{black}    98.81$\dag$}\\
{\color{black}FDNAS(on device-8)}          &3.38      &346.64     &59.0  &{\color{black}    99.22$\dag$}\\
{\color{black}FDNAS(on device-9)}          &3.38      &346.64     &59.0  &{\color{black}    99.60$\dag$}\\
FDNAS(on all devices)          &3.38      &346.64     &59.0  &  78.75$^*$/97.25$\dag$\\ \hline

{\color{black}CFDNAS-G(on device-0)}        &3.33       &318.44     &3.53           &{\color{black} 73.56$^*$/98.39$\dag$}\\
{\color{black}CFDNAS-G(on device-1)}        &3.33       &318.44     &3.53   &{\color{black} 73.80$^*$/99.02$\dag$}\\
{\color{black}CFDNAS-G(on device-2)}        &3.33       &318.44     &3.53               &{\color{black} 73.44$^*$/98.38$\dag$}\\
{\color{black}CFDNAS-G(mean on device-0, 1, 2)}        &3.33       &318.44     &3.53   &{\color{black} 73.60$^*$/98.93$\dag$}\\ \hline

{\color{black}CFDNAS-C(on device-3)}       &2.03       &169.35       &3.46                &{\color{black} 72.01$^*$/93.21$\dag$}\\
{\color{black}CFDNAS-C(on device-4)}      &2.03       &169.35       &3.46   &{\color{black} 70.67$^*$/92.73$\dag$}\\
{\color{black}CFDNAS-C(on device-5)}       &2.03       &169.35       &3.46               &{\color{black} 71.19$^*$/93.01$\dag$}\\
{\color{black}CFDNAS-C(mean on device-3, 4, 5)}       &2.03       &169.35       &3.46       &{\color{black} 71.29$^*$/93.01$\dag$}\\

\bottomrule
\end{tabular}
\label{ablation2}
\end{table*}

\subsubsection{Effectiveness of CFDNAS}
From the Table~\ref{ablation1}, we compare CFDNAS with FDNAS and MobileNetV2.
Both CFDNAS-G (GPU platform) and CFDNAS-C (CPU platform) are trained for 25 epochs, each based on the inherited FDNAS SuperNet.
Both tailored CFDNAS-C and CFDNAS-G normal nets have less FLOPs than FDNAS. They also have lower GPU/CPU inference latency than both FDNAS normal net and the hand-crafted MobileNetV2.
We then study the accuracy improvement introduced by CFDNAS in Table~\ref{ablation3}.
Benefiting from the clustering approach, for the GPU group (including  device-0, 1, and 2), CFDNAS-G searches out the tailored model to achieve $98.93\%$ accuracy. It is more accurate than the original FDNAS on devices of the same GPU group and requires only 3.53 GPU hours of SuperNet adaptation, which is a negligible additional cost.
The ``naive-CFDNAS'' has no inheritance parameters ($\mathbf w$ and $\pmb \alpha$) and no SuperNet's ``meta-test'' adaptation. As a result, the convergence time of naive-CFDNAS takes 18 GPU hours, which is 5.1 times slower than that of CFDNAS. In the same GPU group, the tailored CFDNAS-G normal net is $1.26\%$ more accurate than naive-CFDNAS, while its FLOPs are smaller than naive-CFDNAS.
For the CPU group (including device-3, 4, and 5), the tailored CFDNAS-C normal net also outperforms FDNAS and naive-CFDNAS-C in terms of accuracy and FLOPs.
Data of CPU group is more complicated than others, but the tailored CFDNAS-C normal net is still more accurate than naive-CFDNAS-C and more stable than FDNAS.

{\color{black} To show the importance of the clustering step, we evaluate the efficiency of the normal nets directly  derived from the FDNAS SuperNet (with no clustering), see ``Personalized-FDNAS'' in Table VI. It can be observed that the proposed scheme with clustering mechanism achieves more efficient normal net for each device. This is reasonable since the clustering step allows the knowledge transfer across the devices with similar data distributions and hardware.}

Compared to FDNAS, CFDNAS has an extra meta-adaption mechanism, and tailored CFDNAS normal nets get a better trade-off between accuracy and latency.
Compared to naive-CFDNAS, CFDNAS inherits external ``meta-trained" knowledge (FDNAS SuperNet's $\mathbf w$ and $\pmb \alpha$, which is fully trained using data of all devices), and thus gets better accuracy and less search time cost than naive-CFDNAS. 

\subsubsection{Contributions of federated mechanism}
{\color{black}
In Table~\ref{ablation2}, we show the effects of using a traditional DNAS with well-collected data (including all data) and a single-device DNAS (including only local devices' data, with no federated training or other devices' data).
}
DNAS searches the model directly on collected data from devices but requires data collection in advance, and it achieves a centralized training accuracy of $96.71\%$ in a traditional manner. However, FDNAS still has higher local accuracy than DNAS while protecting data privacy.
Besides, FDNAS has lower FLOPs than single-device conventional DNAS results. Federated averaged accuracy is $78.75\%$, and local accuracy is $97.25\%$, which is $2.94\%$ higher than single-device local average accuracy. Thanks to the federated mechanism, FDNAS can use data from a wide range of devices to search for more efficient models. At the same time, it trains models with higher accuracy. Privacy protection and efficiency will be beneficial for the social impacts and effectiveness of practical machine learning deployments.

{\color{black} {\it Remark 3}: For CFDNAS-G, the mean value of federated average accuracies  on device-0,1,2 ($73.60\%$) is higher than that of FDNAS ($73.50\%$) on device-0,1,2;  for CFDNAS-C, the mean value of federated average accuracies  on device-3,4,5 ($71.29\%$) is close to that of  FDNAS on device-3,4,5 ($71.90\%$) \textit{when keeping devices and dataset (including training set and test set) consistent.} Since we use hardware-aware latency loss to regularize the CFDNAS for more light-weight neural networks, the parameter amount and inference cost (FLOPs, tagged as $\times+$ in tables) of CFDNAS normal nets both decrease. This model size's reduction causes the CFDNAS-C's federated average accuracy to be $71.29\%$ on device-3,4,5, which is slightly lower than FDNAS's federated average accuracy $71.90\%$ on device-3,4,5. However, CFDNAS-C's FLOPs are 169.35M and the parameter amount is 2.03M, which are much smaller than those of FDNAS, whose parameter amount  is 3.38M and FLOPs are 346.64M. In other word, the proposed CFDNAS-C reduces inference computation by \textbf{50\%}, although with a slightly lower accuracy. This is indeed a trade-off between federated average accuracy and model size, especially in AIoT applications. }

{\color{black} {\it Remark 4}: To ensure a fair comparison, further explanation on the test sets for various methods are given as follows. Firstly, the test sets of DNAS device-0,  DNAS device-1, and  DNAS device-2,  are the same as CDFNAS device-0, CDFNAS device-1, and CDFNAS device-2, respectively, so as those of other devices. Secondly, when comparing with DNAS, we train DNAS in a traditional centralized setting. When examining DNAS (on device-i)'s performance, the SuperNet training of DNAS (on device-i) is only optimized on device-i's dataset, and the derived normal net is only trained on device-i's dataset too. So each DNAS (on device-i) is optimized on local dataset and tested on local dataset instead of training or testing on all devices or CIFAR-10. }

\subsubsection{Contributions of direct search}
For a fair comparison, we use MobileNetV2, which is similarly composed of MBconv blocks, as a pre-defined hand-crafted neural architecture trained in FedAvg.
We present the FedAvg results for FDNAS and MobileNetV2 in Table~\ref{ablation1}. 
For FDNAS, the federated averaged accuracy and local accuracy are $10.3\%$ and $0.74\%$ higher than MobileNetV2, respectively.
Although both have similar inference latency on GPU, FDNAS can be faster than MobileNetV2 on the CPU platform. So compared to the FDNAS, MobileNetV2 is not optimal for various devices in federated scenarios.
The FDNAS search neural architecture automatically from the data is superior to hand-crafted models. It concludes that NAS approaches can significantly outperform the manually designed models in federated learning.

\section{Conclusions and Future Research}
\label{sec:con}
In this paper, we have proposed FDNAS, a privacy-preserving neural architecture search scheme under the framework of federated learning. 
Unlike traditional federated learning approaches, the proposed FDNAS seeks the complete neural architecture from both the data and the hardware latency tables of devices automatically.  Extensive numerical results have demonstrated that the proposed FDNAS greatly increases the accuracy of the model while reduces the computational cost, compared to pre-defined and hand-crafted neural architectures in federated learning.
On the other hand, the proposed FDNAS model achieves state-of-the-art results on ImageNet in a mobile setting, demonstrating the transferability of FDNAS. Moreover, inspired by meta-learning, CFDNAS, an extension to FDNAS, can discover diverse \textit{high-accuracy} and \textit{low-latency} models adapted from a SuperNet of FDNAS at a much low training cost for different AIoT platforms. In future work, we will extend  FDNAS for different tasks such as object detection, semantic segmentation, model compression, etc., as well as the scenarios involving more complex datasets and tinier AIoT devices. {\color{black} Also, we will investigate more efficient training algorithms for federated NAS, in order to accelerate its wide use in practical AIoT devices. Finally, it is worthwhile focusing on analyzing the local NAS workload and including strategies to reduce it.}


%





\ifCLASSOPTIONcaptionsoff
  \newpage
\fi



\bibliographystyle{IEEEtran}
%



%
\bibliography{ref.bib}

\begin{thebibliography}{10}
\providecommand{\url}[1]{#1}
\csname url@samestyle\endcsname
\providecommand{\newblock}{\relax}
\providecommand{\bibinfo}[2]{#2}
\providecommand{\BIBentrySTDinterwordspacing}{\spaceskip=0pt\relax}
\providecommand{\BIBentryALTinterwordstretchfactor}{4}
\providecommand{\BIBentryALTinterwordspacing}{\spaceskip=\fontdimen2\font plus
\BIBentryALTinterwordstretchfactor\fontdimen3\font minus
  \fontdimen4\font\relax}
\providecommand{\BIBforeignlanguage}[2]{{%
\expandafter\ifx\csname l@#1\endcsname\relax
\typeout{** WARNING: IEEEtran.bst: No hyphenation pattern has been}%
\typeout{** loaded for the language `#1'. Using the pattern for}%
\typeout{** the default language instead.}%
\else
\language=\csname l@#1\endcsname
\fi
#2}}
\providecommand{\BIBdecl}{\relax}
\BIBdecl

\bibitem{krizhevsky2012imagenet}
A.~Krizhevsky, I.~Sutskever, and G.~E. Hinton, ``Imagenet classification with
  deep convolutional neural networks,'' in \emph{Advances in neural information
  processing systems}, 2012, pp. 1097--1105.

\bibitem{simonyan2014very}
K.~Simonyan and A.~Zisserman, ``Very deep convolutional networks for
  large-scale image recognition,'' in \emph{International Conference on
  Learning Representations}, 2015.

\bibitem{he2016deep}
K.~He, X.~Zhang, S.~Ren, and J.~Sun, ``Deep residual learning for image
  recognition,'' in \emph{Proceedings of the IEEE conference on computer vision
  and pattern recognition}, 2016, pp. 770--778.

\bibitem{hu2018squeeze}
J.~Hu, L.~Shen, and G.~Sun, ``Squeeze-and-excitation networks,'' in
  \emph{Proceedings of the IEEE conference on computer vision and pattern
  recognition}, 2018, pp. 7132--7141.

\bibitem{howard2019searching}
A.~Howard, M.~Sandler, G.~Chu, L.-C. Chen, B.~Chen, M.~Tan, W.~Wang, Y.~Zhu,
  R.~Pang, V.~Vasudevan \emph{et~al.}, ``Searching for mobilenetv3,'' in
  \emph{Proceedings of the IEEE International Conference on Computer Vision},
  2019, pp. 1314--1324.

\bibitem{redmon2018yolov3}
J.~Redmon and A.~Farhadi, ``Yolov3: An incremental improvement,'' \emph{arXiv
  preprint arXiv:1804.02767}, 2018.

\bibitem{ren2015faster}
S.~Ren, K.~He, R.~Girshick, and J.~Sun, ``Faster r-cnn: Towards real-time
  object detection with region proposal networks,'' in \emph{Advances in neural
  information processing systems}, 2015, pp. 91--99.

\bibitem{long2015fully}
J.~Long, E.~Shelhamer, and T.~Darrell, ``Fully convolutional networks for
  semantic segmentation,'' in \emph{Proceedings of the IEEE conference on
  computer vision and pattern recognition}, 2015, pp. 3431--3440.

\bibitem{ji2020mcunet}
J.~Lin, W.-M. Chen, J.~Cohn, C.~Gan, and S.~Han, ``Mcunet: Tiny deep learning
  on iot devices,'' in \emph{Annual Conference on Neural Information Processing
  Systems (NeurIPS)}, 2020.

\bibitem{8693826}
S.~{Alyamkin}, M.~{Ardi}, A.~C. {Berg}, A.~{Brighton}, B.~{Chen}, Y.~{Chen},
  H.~{Cheng}, Z.~{Fan}, C.~{Feng}, B.~{Fu}, K.~{Gauen}, A.~{Goel},
  A.~{Goncharenko}, X.~{Guo}, S.~{Ha}, A.~{Howard}, X.~{Hu}, Y.~{Huang},
  D.~{Kang}, J.~{Kim}, J.~G. {Ko}, A.~{Kondratyev}, J.~{Lee}, S.~{Lee},
  S.~{Lee}, Z.~{Li}, Z.~{Liang}, J.~{Liu}, X.~{Liu}, Y.~{Lu}, Y.~{Lu},
  D.~{Malik}, H.~H. {Nguyen}, E.~{Park}, D.~{Repin}, L.~{Shen}, T.~{Sheng},
  F.~{Sun}, D.~{Svitov}, G.~K. {Thiruvathukal}, B.~{Zhang}, J.~{Zhang},
  X.~{Zhang}, and S.~{Zhuo}, ``Low-power computer vision: Status, challenges,
  and opportunities,'' \emph{IEEE Journal on Emerging and Selected Topics in
  Circuits and Systems}, vol.~9, no.~2, pp. 411--421, 2019.

\bibitem{wang2020scaled}
C.-Y. Wang, A.~Bochkovskiy, and H.-Y.~M. Liao, ``Scaled-yolov4: Scaling cross
  stage partial network,'' \emph{arXiv preprint arXiv:2011.08036}, 2020.

\bibitem{pmlr-v54-mcmahan17a}
B.~McMahan, E.~Moore, D.~Ramage, S.~Hampson, and B.~A. y~Arcas,
  ``Communication-efficient learning of deep networks from decentralized
  data,'' in \emph{Artificial Intelligence and Statistics}, 2017, pp.
  1273--1282.

\bibitem{truex2019hybrid}
S.~Truex, N.~Baracaldo, A.~Anwar, T.~Steinke, H.~Ludwig, R.~Zhang, and Y.~Zhou,
  ``A hybrid approach to privacy-preserving federated learning,'' in
  \emph{Proceedings of the 12th ACM Workshop on Artificial Intelligence and
  Security}, 2019, pp. 1--11.

\bibitem{li2020on}
X.~Li, K.~Huang, W.~Yang, S.~Wang, and Z.~Zhang, ``On the convergence of fedavg
  on non-iid data,'' in \emph{International Conference on Learning
  Representations}, 2020.

\bibitem{liu2018darts}
H.~Liu, K.~Simonyan, and Y.~Yang, ``{DARTS}: Differentiable architecture
  search,'' in \emph{International Conference on Learning Representations},
  2019.

\bibitem{cai2018proxylessnas}
H.~Cai, L.~Zhu, and S.~Han, ``Proxyless{NAS}: Direct neural architecture search
  on target task and hardware,'' in \emph{International Conference on Learning
  Representations}, 2019.

\bibitem{xu2020pcdarts}
Y.~Xu, L.~Xie, X.~Zhang, X.~Chen, G.-J. Qi, Q.~Tian, and H.~Xiong, ``Pc-darts:
  Partial channel connections for memory-efficient architecture search,'' in
  \emph{International Conference on Learning Representations}, 2020.

\bibitem{he2020milenas}
C.~He, H.~Ye, L.~Shen, and T.~Zhang, ``Milenas: Efficient neural architecture
  search via mixed-level reformulation,'' in \emph{Proceedings of the IEEE/CVF
  Conference on Computer Vision and Pattern Recognition}, 2020, pp.
  11\,993--12\,002.

\bibitem{yang2019evaluation}
A.~Yang, P.~M. Esperança, and F.~M. Carlucci, ``Nas evaluation is
  frustratingly hard,'' in \emph{International Conference on Learning
  Representations}, 2020.

\bibitem{8889996}
F.~Sattler, S.~Wiedemann, K.-R. M{\"u}ller, and W.~Samek, ``Robust and
  communication-efficient federated learning from non-iid data,'' \emph{IEEE
  transactions on neural networks and learning systems}, 2019.

\bibitem{chen2019closerfewshot}
W.-Y. Chen, Y.-C. Liu, Z.~Kira, Y.-C. Wang, and J.-B. Huang, ``A closer look at
  few-shot classification,'' in \emph{International Conference on Learning
  Representations}, 2019.

\bibitem{he2020fednas}
C.~He, M.~Annavaram, and S.~Avestimehr, ``Fednas: Federated deep learning via
  neural architecture search,'' \emph{arXiv preprint arXiv:2004.08546}, 2020.

\bibitem{cifar10}
A.~Krizhevsky, G.~Hinton \emph{et~al.}, ``Learning multiple layers of features
  from tiny images,'' \emph{University of Toronto}, 2009.

\bibitem{5206848}
J.~Deng, W.~Dong, R.~Socher, L.-J. Li, K.~Li, and L.~Fei-Fei, ``Imagenet: A
  large-scale hierarchical image database,'' in \emph{2009 IEEE conference on
  computer vision and pattern recognition}, 2009, pp. 248--255.

\bibitem{caldas2018leaf}
S.~Caldas, S.~M.~K. Duddu, P.~Wu, T.~Li, J.~Kone{\v{c}}n{\`y}, H.~B. McMahan,
  V.~Smith, and A.~Talwalkar, ``Leaf: A benchmark for federated settings,''
  \emph{arXiv preprint arXiv:1812.01097}, 2018.

\bibitem{he2020fedml}
C.~He, S.~Li, J.~So, X.~Zeng, M.~Zhang, H.~Wang, X.~Wang, P.~Vepakomma,
  A.~Singh, H.~Qiu \emph{et~al.}, ``Fedml: A research library and benchmark for
  federated machine learning,'' \emph{arXiv preprint arXiv:2007.13518}, 2020.

\bibitem{sandler2018mobilenetv2}
M.~Sandler, A.~Howard, M.~Zhu, A.~Zhmoginov, and L.-C. Chen, ``Mobilenetv2:
  Inverted residuals and linear bottlenecks,'' in \emph{Proceedings of the IEEE
  conference on computer vision and pattern recognition}, 2018, pp. 4510--4520.

\bibitem{brock2018smash}
A.~Brock, T.~Lim, J.~Ritchie, and N.~Weston, ``{SMASH}: One-shot model
  architecture search through hypernetworks,'' in \emph{International
  Conference on Learning Representations}, 2018.

\bibitem{pham2018efficient}
H.~Pham, M.~Guan, B.~Zoph, Q.~Le, and J.~Dean, ``Efficient neural architecture
  search via parameters sharing,'' in \emph{International Conference on Machine
  Learning}, ser. Proceedings of Machine Learning Research, vol.~80, 10--15 Jul
  2018, pp. 4095--4104.

\bibitem{wu2019fbnet}
B.~Wu, X.~Dai, P.~Zhang, Y.~Wang, F.~Sun, Y.~Wu, Y.~Tian, P.~Vajda, Y.~Jia, and
  K.~Keutzer, ``Fbnet: Hardware-aware efficient convnet design via
  differentiable neural architecture search,'' in \emph{Proceedings of the IEEE
  Conference on Computer Vision and Pattern Recognition}, 2019, pp.
  10\,734--10\,742.

\bibitem{wan2020fbnetv2}
A.~Wan, X.~Dai, P.~Zhang, Z.~He, Y.~Tian, S.~Xie, B.~Wu, M.~Yu, T.~Xu, K.~Chen
  \emph{et~al.}, ``Fbnetv2: Differentiable neural architecture search for
  spatial and channel dimensions,'' in \emph{Proceedings of the IEEE/CVF
  Conference on Computer Vision and Pattern Recognition}, 2020, pp.
  12\,965--12\,974.

\bibitem{li2019federated}
T.~Li, A.~K. Sahu, A.~Talwalkar, and V.~Smith, ``Federated learning:
  Challenges, methods, and future directions,'' \emph{IEEE Signal Processing
  Magazine}, vol.~37, no.~3, pp. 50--60, 2020.

\bibitem{konevcny2016federated}
J.~Kone{\v{c}}n{\`y}, H.~B. McMahan, F.~X. Yu, P.~Richt{\'a}rik, A.~T. Suresh,
  and D.~Bacon, ``Federated learning: Strategies for improving communication
  efficiency,'' \emph{arXiv preprint arXiv:1610.05492}, 2016.

\bibitem{45672}
A.~T. Suresh, X.~Y. Felix, S.~Kumar, and H.~B. McMahan, ``Distributed mean
  estimation with limited communication,'' in \emph{International Conference on
  Machine Learning}, 2017, pp. 3329--3337.

\bibitem{fedpaq19}
A.~Reisizadeh, A.~Mokhtari, H.~Hassani, A.~Jadbabaie, and R.~Pedarsani,
  ``Fedpaq: A communication-efficient federated learning method with periodic
  averaging and quantization,'' in \emph{International Conference on Artificial
  Intelligence and Statistics}, 2020, pp. 2021--2031.

\bibitem{9272666}
G.~{Zhu}, Y.~{Du}, D.~{Gündüz}, and K.~{Huang}, ``One-bit over-the-air
  aggregation for communication-efficient federated edge learning: Design and
  convergence analysis,'' \emph{IEEE Transactions on Wireless Communications},
  vol.~20, no.~3, pp. 2120--2135, 2021.

\bibitem{zhao2018federated}
Y.~Zhao, M.~Li, L.~Lai, N.~Suda, D.~Civin, and V.~Chandra, ``Federated learning
  with non-iid data,'' \emph{arXiv preprint arXiv:1806.00582}, 2018.

\bibitem{agarwal2018cpsgd}
N.~Agarwal, A.~T. Suresh, F.~X.~X. Yu, S.~Kumar, and B.~McMahan, ``cpsgd:
  Communication-efficient and differentially-private distributed sgd,'' in
  \emph{Advances in Neural Information Processing Systems}, 2018, pp.
  7564--7575.

\bibitem{article17eth}
R.~C. Geyer, T.~Klein, and M.~Nabi, ``Differentially private federated
  learning: A client level perspective,'' \emph{arXiv preprint
  arXiv:1712.07557}, 2017.

\bibitem{9069945}
K.~Wei, J.~Li, M.~Ding, C.~Ma, H.~H. Yang, F.~Farokhi, S.~Jin, T.~Q. Quek, and
  H.~V. Poor, ``Federated learning with differential privacy: Algorithms and
  performance analysis,'' \emph{IEEE Transactions on Information Forensics and
  Security}, 2020.

\bibitem{cai2020once}
H.~Cai, C.~Gan, T.~Wang, Z.~Zhang, and S.~Han, ``Once for all: Train one
  network and specialize it for efficient deployment,'' in \emph{International
  Conference on Learning Representations}, 2020.

\bibitem{courbariaux2015binaryconnect}
M.~Courbariaux, Y.~Bengio, and J.-P. David, ``Binaryconnect: Training deep
  neural networks with binary weights during propagations,'' in \emph{Advances
  in neural information processing systems}, 2015, pp. 3123--3131.

\bibitem{finn2017model}
C.~Finn, P.~Abbeel, and S.~Levine, ``Model-agnostic meta-learning for fast
  adaptation of deep networks,'' in \emph{International Conference on Machine
  Learning}.\hskip 1em plus 0.5em minus 0.4em\relax PMLR, 2017, pp. 1126--1135.

\bibitem{nichol2018first}
A.~Nichol, J.~Achiam, and J.~Schulman, ``On first-order meta-learning
  algorithms,'' \emph{arXiv preprint arXiv:1803.02999}, 2018.

\bibitem{densenet}
G.~Huang, Z.~Liu, L.~Van Der~Maaten, and K.~Q. Weinberger, ``Densely connected
  convolutional networks,'' in \emph{Proceedings of the IEEE conference on
  computer vision and pattern recognition}, 2017, pp. 4700--4708.

\bibitem{zoph2018learning}
B.~Zoph, V.~Vasudevan, J.~Shlens, and Q.~V. Le, ``Learning transferable
  architectures for scalable image recognition,'' in \emph{Proceedings of the
  IEEE conference on computer vision and pattern recognition}, 2018, pp.
  8697--8710.

\bibitem{amoebanet}
E.~Real, A.~Aggarwal, Y.~Huang, and Q.~V. Le, ``Regularized evolution for image
  classifier architecture search,'' in \emph{Proceedings of the aaai conference
  on artificial intelligence}, vol.~33, 2019, pp. 4780--4789.

\bibitem{liu2018hierarchical}
H.~Liu, K.~Simonyan, O.~Vinyals, C.~Fernando, and K.~Kavukcuoglu,
  ``Hierarchical representations for efficient architecture search,'' in
  \emph{International Conference on Learning Representations}, 2018.

\bibitem{liu2018progressive}
C.~Liu, B.~Zoph, M.~Neumann, J.~Shlens, W.~Hua, L.-J. Li, L.~Fei-Fei,
  A.~Yuille, J.~Huang, and K.~Murphy, ``Progressive neural architecture
  search,'' in \emph{Proceedings of the European Conference on Computer
  Vision}, 2018, pp. 19--34.

\bibitem{xie2018snas}
S.~Xie, H.~Zheng, C.~Liu, and L.~Lin, ``{SNAS}: stochastic neural architecture
  search,'' in \emph{International Conference on Learning Representations},
  2019.

\bibitem{chen2019progressive}
X.~Chen, L.~Xie, J.~Wu, and Q.~Tian, ``Progressive differentiable architecture
  search: Bridging the depth gap between search and evaluation,'' in
  \emph{Proceedings of the IEEE International Conference on Computer Vision},
  2019, pp. 1294--1303.

\bibitem{dong2019searching}
X.~Dong and Y.~Yang, ``Searching for a robust neural architecture in four gpu
  hours,'' in \emph{Proceedings of the IEEE Conference on Computer Vision and
  Pattern Recognition}, 2019, pp. 1761--1770.

\bibitem{paszke2019pytorch}
A.~Paszke, S.~Gross, F.~Massa, A.~Lerer, J.~Bradbury, G.~Chanan, T.~Killeen,
  Z.~Lin, N.~Gimelshein, L.~Antiga \emph{et~al.}, ``Pytorch: An imperative
  style, high-performance deep learning library,'' in \emph{Advances in neural
  information processing systems}, 2019, pp. 8026--8037.

\bibitem{howard2017mobilenets}
A.~G. Howard, M.~Zhu, B.~Chen, D.~Kalenichenko, W.~Wang, T.~Weyand,
  M.~Andreetto, and H.~Adam, ``Mobilenets: Efficient convolutional neural
  networks for mobile vision applications,'' \emph{arXiv preprint
  arXiv:1704.04861}, 2017.

\bibitem{ma2018shufflenet}
N.~Ma, X.~Zhang, H.-T. Zheng, and J.~Sun, ``Shufflenet v2: Practical guidelines
  for efficient cnn architecture design,'' in \emph{Proceedings of the European
  conference on computer vision}, 2018, pp. 116--131.

\bibitem{tan2019mnasnet}
M.~Tan, B.~Chen, R.~Pang, V.~Vasudevan, M.~Sandler, A.~Howard, and Q.~V. Le,
  ``Mnasnet: Platform-aware neural architecture search for mobile,'' in
  \emph{Proceedings of the IEEE Conference on Computer Vision and Pattern
  Recognition}, 2019, pp. 2820--2828.

\bibitem{xie2019exploring}
S.~Xie, A.~Kirillov, R.~Girshick, and K.~He, ``Exploring randomly wired neural
  networks for image recognition,'' in \emph{Proceedings of the IEEE
  International Conference on Computer Vision}, 2019, pp. 1284--1293.

\end{thebibliography}

 \begin{IEEEbiography}
 [{\includegraphics[width=1in,height=1.25in,clip,keepaspectratio]{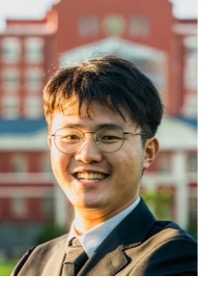}}]  {Chunhui Zhang}
 received the B.Eng. degree from Northeastern University at Qinhuangdao in 2021. Currently, he is a first year Ph.D. student with the Michtom School of Computer Science at Brandeis University. His research interests are in efficient machine learning, representation learning, and graph neural network. 
 \end{IEEEbiography}

 \begin{IEEEbiography}
 [{\includegraphics[width=1in,height=1.25in,clip,keepaspectratio]{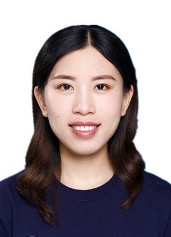}}]  {Xiaoming Yuan} received her B.E. degree in Electronics and Information Engineering from Henan Polytechnic University, China, in 2012. She received the Ph.D. degree in Communication and Information System from Xidian University, China, in 2018. She is now a lecturer of Qinhuangdao Branch Campus, Northeastern University, China, from 2018. Her research interests include cloud/edge computing, artificial intelligence, resource management for the Internet of Vehicles and the Internet of Medical Things.
 \end{IEEEbiography}

  \begin{IEEEbiography}
 [{\includegraphics[width=1in,height=1.25in,clip,keepaspectratio]{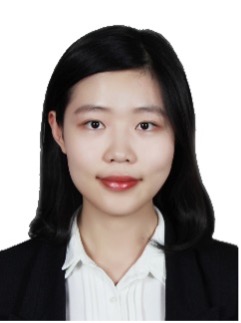}}]  {Qianyun Zhang} received her B.Sc. degree from Beijing University of Posts and Telecommunications, China, in 2014, and her Ph.D. degree from Queen Mary University of London, United Kingdom, in 2018. She is currently an Associate Professor with the School of Cyber Science and Technology, Beihang University, Beijing, China. Her research interests include wireless network security, intelligent sensing and recognition, and novel antenna designs.
 \end{IEEEbiography}

   \begin{IEEEbiography}
 [{\includegraphics[width=1in,height=1.25in,clip,keepaspectratio]{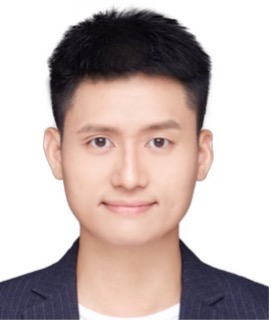}}]  {Guangxu Zhu} received the B.S. and M.S. degrees in electronic and electrical engineering from Zhejiang University and the Ph.D. degree in electronic and electrical engineering from The University of Hong Kong. He is currently a Research Scientist with the Shenzhen Research Institute of Big Data. His research interests include edge intelligence, distributed machine learning, and 5G technologies, such as massive MIMO, mmWave communication, and wirelessly powered communication. He was a recipient of the Hong Kong Postgraduate Fellowship (HKPF), the Outstanding Ph.D. Thesis Award from HKU, and the Best Paper Award from WCSP 2013. He served as a Co-Chair for the "MAC and cross-layer design" track in IEEE PIMRC 2021. 
 \end{IEEEbiography}

   \begin{IEEEbiography}
 [{\includegraphics[width=1in,height=1.25in,clip,keepaspectratio]{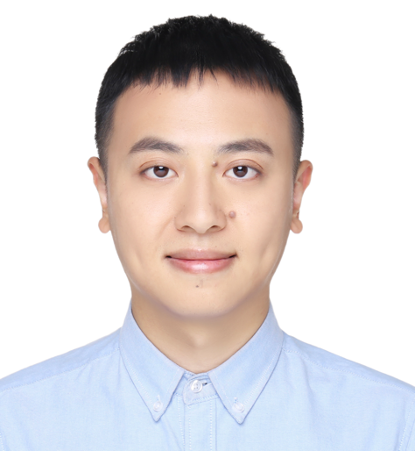}}]  {Lei Cheng} is an Assistant Professor (ZJU Young Professor) in the College of Information Science and Electronic Engineering at Zhejiang University, Hangzhou, China. He received the B.Eng. degree from Zhejiang University in 2013, and the Ph.D. degree from the University of Hong Kong in 2018. He was a research scientist in Shenzhen Research Institute of Big Data from 2018 to 2021. His research interests are in Bayesian machine learning for tensor data analytics, and interpretable machine learning for information systems.
 \end{IEEEbiography}

    \begin{IEEEbiography}
 [{\includegraphics[width=1in,height=1.25in,clip,keepaspectratio]{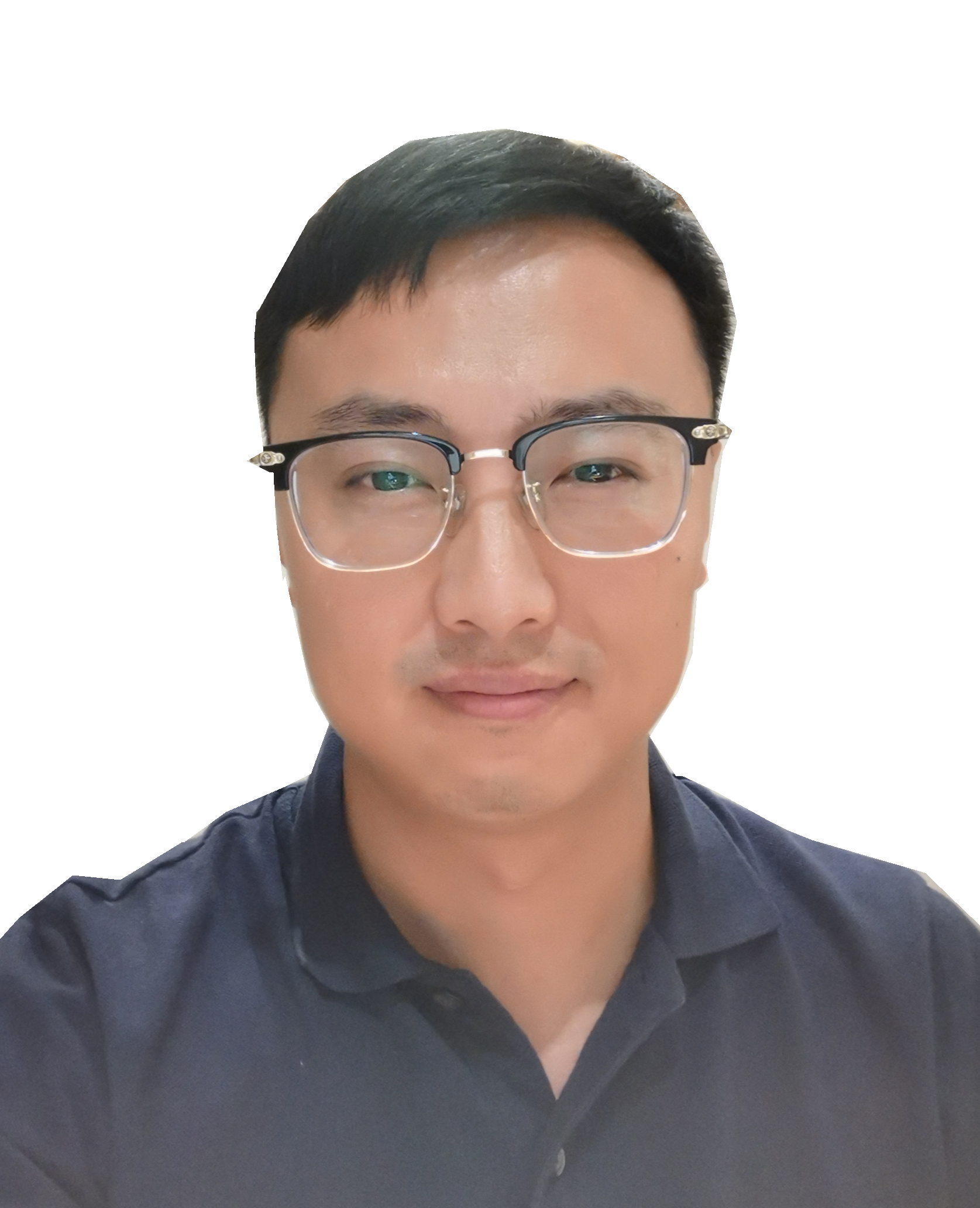}}]  {Ning Zhang}  is an Associate Professor and Canada Research Chair in the Department of Electrical and Computer Engineering at University of Windsor, Canada. He received the Ph.D degree in Electrical and Computer Engineering from University of Waterloo, Canada, in 2015. After that, he was a postdoc research fellow at University of Waterloo and University of Toronto, respectively. His research interests include connected vehicles, mobile edge computing, wireless networking, and machine learning. He is a Highly Cited Researcher. He serves as an Associate Editor of IEEE Internet of Things Journal and IEEE Transactions on Cognitive Communications and Networking; and a Guest Editor of several international journals, such as IEEE Wireless Communications, IEEE Transactions on Industrial Informatics, and IEEE Transactions on Intelligent Transportation Systems. He also serves/served as a TPC chair for IEEE VTC 2021 and IEEE SAGC 2020, a general chair for IEEE SAGC 2021, a track chair for several international conferences and workshops. He received 8 Best Paper Awards from conferences and journals, such as IEEE Globecom and IEEE ICC.  
 \end{IEEEbiography}
 






\end{document}